\pdfoutput=1
\documentclass[11pt]{article}

\usepackage{acl}
\usepackage{times}
\usepackage{latexsym}

\usepackage[T1]{fontenc}
\usepackage[utf8]{inputenc}

\usepackage{microtype}

\usepackage{inconsolata}

\usepackage{microtype}

\usepackage{makecell}
\usepackage{setspace}
\usepackage{lipsum}
\usepackage{tabularx}
\usepackage{amsmath}
\usepackage{graphicx,subcaption}
\usepackage{multirow}
\usepackage{booktabs}
\usepackage[noend]{algpseudocode}
\usepackage{enumitem}
\usepackage{placeins}
\usepackage{cancel} % add by wjn
\usepackage{CJKutf8}
\usepackage[ruled,linesnumbered]{algorithm2e}
\usepackage{soul}
\definecolor{mylightgreen}{RGB}{144,238,144}

\title{Aligning Large Language Models to a Domain-specific Graph Database for NL2GQL}
\author{
     Yuanyuan Liang$^1$, Keren Tan$^1$, Tingyu Xie$^2$,  Wenbiao Tao$^1$ ,  Siyuan Wang$^1$\\
      \textbf{Yunshi Lan$^1$\thanks{*Corresponding author},Weining Qian$^1$}\\
  $^1$ East China Normal University, $^2$ Zhejiang University\\
  	\{leonyuany,tankeren1020,wbtao,sywang\}@stu.ecnu.edu.cn\\
        tingyuxie@zju.edu.cn, \{yslan,wnqian\}@dase.ecnu.edu.cn 
   }

\begin{document}
\maketitle
\begin{abstract}
Graph Databases (Graph DB) find extensive application across diverse domains such as finance, social networks, and medicine. 
Yet, the translation of Natural Language (NL) into the Graph Query Language (GQL), referred to as NL2GQL, poses significant challenges owing to its intricate and specialized nature.
Some approaches have sought to utilize Large Language Models (LLMs) to address analogous tasks like text2SQL. 
Nonetheless, in the realm of NL2GQL tasks tailored to a particular domain, the absence of domain-specific NL-GQL data pairs adds complexity to aligning LLMs with the graph DB.
To tackle this challenge, we present a well-defined pipeline. Initially, we utilize ChatGPT to generate NL-GQL data pairs, leveraging the provided graph DB with self-instruction. Subsequently, we employ the generated data to fine-tune LLMs, ensuring alignment between LLMs and the graph DB. 
Moreover, we find the importance of relevant schema in efficiently generating accurate GQLs. Thus, we introduce a method to extract relevant schema as the input context.
We evaluate our method using two carefully constructed datasets derived from graph DBs in the finance and medicine domains, named FinGQL and MediGQL.
Experimental results reveal that our approach significantly outperforms a set of baseline methods, with improvements of 5.90 and 6.36 absolute points on EM, and 6.00 and 7.09 absolute points on EX for FinGQL and MediGQL, respectively.
% \yscomment{specify what detailed results we can achieve, what improvement we obtain.}
% Our method excels in aligning LLMs to domain-specific graph databases, providing a robust solution for NL2GQL problems in specific domains.
\end{abstract}

\section{Introduction}

% \yscomment{Describe the following paragraphs: 1. significance of NL2GQL tasks and their challenges. 2. traditional methods for NL2GQL and their shortcomings (their performance is not ideal). 3. the potential of LLMs on transferring natural language to structured query and identify the unique challenges when we apply these methods to NL2GQL (a. LLMs do not have the knowledge of the specific graph DB. b. LLMs are not familiar with the graph DB languages like neo4j, nabula graph etc,. c. the poor generalization of existing methods is hard to handle the customized graph DB in this task.). 4. we propose a general method to  align LLMs to a given graph DB and what we achieve. 5. summarize our contribution.}

\begin{figure}[]
\centering
\includegraphics[width=0.48\textwidth]{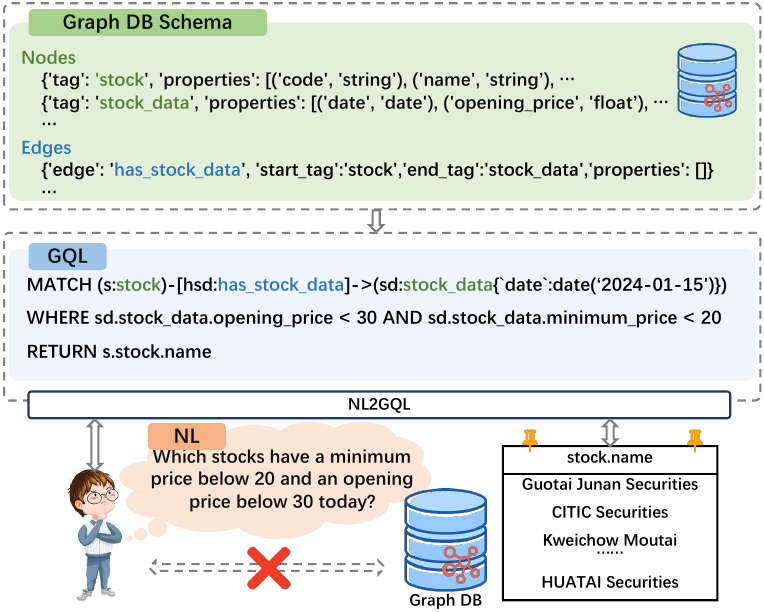}
\caption{An instance of the NL2GQL task in the financial domain, which entails the conversion of NL into GQL tailored for a specific graph DB.}
\label{fig:motivation}
\vspace{-1.5em}
\end{figure}

Presently, leading Graph Databases (Graph DBs) such as Neo4j, NebulaGraph, and JanusGraph are extensively utilized for storing graph data \cite{wang2020empirical,lopes2023scalability}. 
These Graph DBs have broad applications across various domains, including finance, healthcare, social networks, and e-commerce.
However, the nuanced and complex syntax of the Graph Query Language (GQL) presents hurdles for general users endeavoring to convert Natural Language (NL) into GQL.
Figure~\ref{fig:motivation} illustrates a NL2GQL example.
A GQL consists of keywords MATCH, WHERE, and RETURN leading the retrieval functions towards a graph DB.
%A NL2GQL process is illustrated in Figure \ref{fig:motivation}.
%Even though there are analogous tasks to NL2GQL, such as text2SQL~\cite{} and KGQA~\cite{}, the databases they 
% Due to the inherent complexity of natural language understanding, the varied and intricate nature of graph database query languages \cite{monteiro2023experimental}, the diversity in terminology and conventions within domain-specific graph data, and the challenge posed by the scarcity of large-scale, high-quality annotated data needed to train high-performance and generalizable models, there is an additional consideration in practical applications regarding the speed of generating query languages.
% All of these factors together contribute to making NL2GQL an exceedingly challenging task.

Several factors combine to render NL2GQL an exceedingly challenging task. Primarily, the inherent complexity of NL understanding poses initial hurdles due to its intricate nature. Additionally, the intricate syntax of GQL further compounds the difficulty~\cite{monteiro2023experimental}. Furthermore, the diversity of terminology and conventions within domain-specific graph data adds another layer of complexity to the translation process.

% NL2GQL is an exceptionally challenging task for various reason. Firstly, natural language comprehension is inherently difficult, demanding the resolution of semantic ambiguities, adaptation to diverse language styles, and a thorough grasp of the original natural language intent for its transformation into GQL.
% Secondly, the complexity of graph database query languages, involving intricate concepts related to graph structures and node relationships, requires the model to possess a profound understanding of graph databases. Additionally, different graph databases often employ distinct query languages such as Cypher, NGQL, and Gremlin.
% Thirdly, diverse domains may utilize unique terminologies and conventions, necessitating adaptability to ensure effective query language generation within specific domains. 
% Lastly, the scarcity of large-scale, high-quality annotated data makes training a high-performance, generalizable model under low-resource conditions particularly challenging. Moreover, in practical applications, the speed and efficiency of query language generation are pivotal considerations, especially in scenarios demanding real-time responsiveness.

While the NL2GQL task is both highly significant and extremely challenging, recent research on this topic is remarkably scarce. 
The predominant research emphasis is on the conversion of NL into Cypher, which is an efficient GQL aligned to the Nebula Graph.
%Given the widespread adoption of Cypher, the predominant research emphasis is on the conversion of NL into Cypher.
\citet{guo2022spcql} released a general-domain Text-to-Cypher dataset and explored the application of a basic Seq2Seq framework for this task. Nevertheless, the accuracy of their proposed baseline method is notably low, barely surpassing a meager $2\%$. 
\citet{zhou2023r3nl2gql} developed a hybrid approach combining smaller and larger foundation models to improve NL2GQL accuracy in general domain.
Considering the connection between NL2GQL and text2SQL tasks, \citet{zhao_cySpider} developed a SQL2Cypher algorithm capable of mapping SQL queries to Cypher queries. 
However, the relationship between schema items relies heavily on the domain knowledge, which cannot be easily generalized from general domain.
Above approaches are hard to be applied to NL2GQL over a domain-specific graph DB.
%On the one hand, there is lack of annotation of NL-GQL pairs in specific domain for training.
%Leveraging this approach, they constructed the Cy-Spider dataset. Their experimental results indicate that although CodeT5-Base exhibited the highest performance on this dataset, achieving an accuracy of only $71.43\%$.

% \krcomment{In order to draw the following conclusion, maybe elaborate on how poor the performance is, perhaps by adding a numerical value, similar to the mentioned 2\% below.}
% From these two works, it can be observed that utilizing small-scale deep neural network models does not effectively address the NL2GQL task.

In the last year, Large Language Models (LLMs) have transitioned from emerging technologies to widely used applications~\cite{wei2022emergent,zhao2023survey,kaddour2023challenges,xi2023rise}. 
% They have showcased substantial progress in Natural Language Processing (NLP), revealing enhanced capabilities in language comprehension, natural language generation, context understanding, and zero-shot learning. 
Concurrently, a body of research substantiates the robust capacity of LLMs in transforming NL into logical forms~\cite{luo2023chatkbqa,meyer2023llm,li2023flexkbqa} with a lack of training data.
Although these methods have achieved certain results, their methods are tailored to Knowledge Graphs (KGs), which are stored in Resource Description Framework (RDF)~\cite{pan2009resource} and have distinct schema structure to graph DBs, there is still a significant gap for LLMs application in NL2GQL. 
% it is challenging to generalize them to the general NL2GQL task. 
% More importantly, their method fails to address NL2GQL tasks when only a graph DB is provided without any NL-GQL data pairs.
% ~\krcomment{Maybe add a parenthese to clarify what ``structured queries'' means.}

% Despite these advances, applying LLMs to the NL2GQL task encounters significant challenges. LLMs lack specific knowledge about graph databases, which poses a notable limitation. Additionally, these models are unfamiliar with query languages such as Cypher, nGQL, Gremlin, further complicating their integration into tasks related to graph databases. The current methods that leverage LLMs for converting natural language into structured queries also struggle with limited generalization, making it difficult to tailor them effectively for specific graph databases in this task.

% Despite these advances, applying LLMs to the NL2GQL task encounters three significant challenges as follows:

% \noindent \textbf{Lack of knowledge}: LLMs lack specific knowledge about graph databases, which poses a notable limitation.

% \noindent \textbf{Unfamiliar with GQL}: LLMs are unfamiliar with query languages such as Cypher, nGQL, Gremlin, further complicating their integration into tasks related to graph databases.

% \noindent \textbf{Poor generalization}: the current methods that leverage LLMs for converting natural language into structured queries also struggle with limited generalization, making it difficult to tailor them effectively for specific graph databases in this task.

\begin{figure*}[]
\centering
\includegraphics[width=0.95\textwidth]{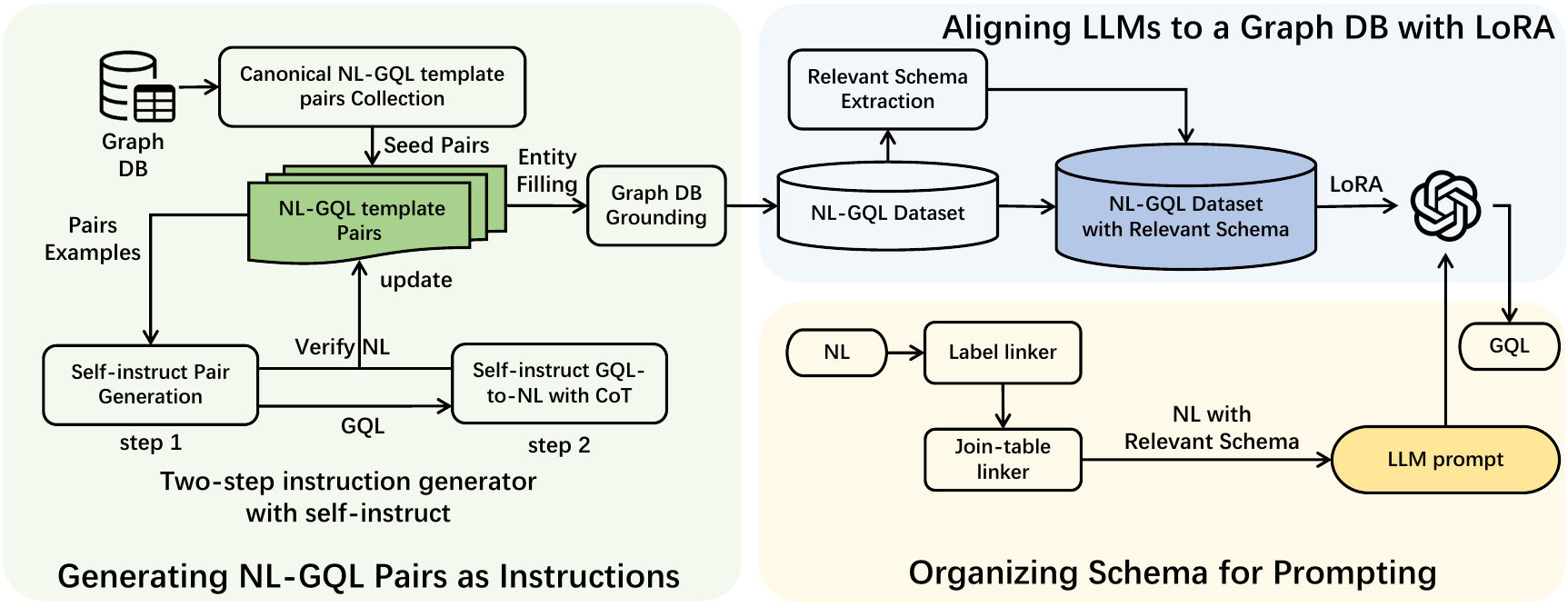}
\caption{Overview of our method. 
}
\vspace{-1.5em}
\label{fig:Overview}
\end{figure*}

In order to improve the performance of NL2GQL tasks on a specific graph DB without any label data, we propose a well-defined pipeline, as depicted in Figure \ref{fig:Overview}. 
%Firstly, we create a data set that includes NL-GQL pairs based on a given Graph DB for fine-tuning LLMs. 
First, we generate NL-GQL pairs as instructions based on a given Graph DB for fine-tuning LLMs.
Specifically, starting from the Graph DB, we collect a set of canonical NL-GQL template pairs. 
Subsequently, we employ a two-step self-instruct~\cite{wang2022self} method to extend these initial pairs, generating a comprehensive collection of NL-GQL template pairs, and further being verified via Chain-of-thought~\cite{wei2022chain}. 
This ensures broad query coverage and diverse expressions while maintaining consistency between NL template and GQL template.
The resulting NL-GQL template pairs undergo graph DB grounding to get complete NL-GQL data pairs.
Secondly, we align a foundation LLM to the graph DB with LoRA~\cite{hu2021lora}.
During inference, we organize the schema that is relevant to NL for prompting the aligned LLMs.
%we integrate a portion of schema that is relevant to NL into the prompt to guide LLMs.
%to ensure that the aligned LLM generates more accurate GQL and accelerates inference, 
We evaluate the performance of our aligned on the graph DBs in finance and medicine domains. 
We select traditional fine-tuned methods and In-context Learning (ICL) methods as baseline methods. 
The experimental results demonstrate that our method outperforms all of the baseline methods by a noticeable margin.
%to ensure that the LLMs is well-aligned with the graph DB, we use parameter-efficient learning technique to fine-tune LLMs using the data set generated from the graph DB.

% It is noteworthy that during the fine-tuning of the LLM, we incorporate the relevant schema by extracting it from the labeled GQL in the training data as an integral part of the input content.
% During the inference phase, to constrain the LLMs with the graph DB schema, we retrieve the relevant schema for the query as the context of the prompt. Subsequently, we instruct the LLM to generate GQL using the identified schema.

We summarize our contributions as follows:
\begin{itemize}
		\setlength{\itemsep}{0pt}
    \item 
     We introduced a pipeline for aligning LLMs with a domain-specific Graph DB, enabling the seamless incorporation of pertinent information from the graph database and its stored data into LLMs.
     % ~\krcomment{Maybe clarify the methods of generating NL-GQL pairs.}
    \item 
     To enhance the accuracy and inference speed of LLM on NL2GQL tasks, we propose a method that extracts a portion of schema from the graph database schema, which is relevant to NL, to serve as input context.
     % ~\st{, using this schema as the prompt context to boost the performance of the NL2GQL task.}
    \item
     Experimental results showcase the superior effectiveness of our proposed approach compared to the baseline.
\end{itemize}

% \begin{figure}[]
% \centering
% \includegraphics[width=0.48\textwidth]{Figures/motivation.pdf}
% % \vspace{-0.5cm}
% \caption{Overview of our method.
% }
% \label{fig:task_def}
% \end{figure}

\section{Related Work}
\subsection{NL2GQL}

% \yscomment{Revise the original Section Text-to-Cypher here, but extend the scope of discussion from text-to-cypher to NL2GQL.}

%NL2GQL is an emerging field of study dedicated to converting natural language queries into formats compatible with graph databases, indicating a growing interest in bridging the gap between human language and database interactions\cite{zhou2023r3nl2gql}. However, with the widespread adoption of Cypher, current research predominantly focuses on NL2Cypher, also known as Text-to-Cypher.
NL2GQL is an emerging task that requires a system to convert a natural language question into formats compatible with graph DB.
The system should have strong capabilities of question understanding and GQL parsing.
Initial endeavors focused on direct translations from natural language to queries, primarily relying on hand-crafted rules~\cite{zhao2022natural}.
More recent studies shift towards data-driven approaches.
By training a sequence-to-sequence model on NL2GQL datasets, a system can learn the transition from NL to GQL well~\cite{guo2022spcql}.
Our study demonstrates innovation in solving NL2SQL over a domain-specific graph DB and aligning LLMs to such a graph DB.
%sequence-to-sequence models for Text-to-Cypher.
%This task involves understanding the graph schema, conducting semantic parsing on questions, and generating queries. It encompasses various aspects of natural language processing technologies, presenting a highly intricate challenge.

%According to \citet{zhao2022natural}, initial endeavors focused on direct translations from natural language to queries, primarily relying on rule-based systems. While these systems provided initial insights, they grappled with the complexity of handling diverse language structures and nuances.
%Recent studies have shifted towards leveraging deep learning techniques to enhance capabilities. The work of \citet{guo2022spcql} introduced a data-driven approach using sequence-to-sequence models for Text-to-Cypher. In the development of a versatile SQL2Cypher algorithm, \citet{zhao_cySpider} mapped SQL queries to a set of Cypher clauses. The resulting Cypher statements were then amalgamated with the original natural language queries to compile a Text-to-Cypher dataset. 

We also notice that text2SQL and KGQA (question answering over knowledge graphs) are analogous tasks to NL2GQL, where the query language are compatible with traditional DBs and knowledge bases, respectively.
Even though prior studies have developed a number of text2SQL as well as KGQA datasets and end-to-end systems~\cite{gao2023text,chang2023prompt,gu2023few,saxena-etal-2020-improving,jiang2022knowledg},
% \yscomment{Please also cite some papers about KGQA using traditional deep learning-based methods.}
they cannot be directly transferred to NL2GQL.
% ~\yscomment{Cite papers}
NL2GQL proves its differences in complex graph modality, variance query types and distinct graph query languages~\cite{zhou2023r3nl2gql}.
While our approach aligns with NL2SQL in terms of the methodological flow, there are substantial disparities in solving NL2GQL over a domain-specific graph DB. % and dataset construction.
%Therefore, it is challenging to solve NL2GQL with a domain-specific graph DB.

%The most analogous task to NL2GQL is Text-to-SQL, as both aim to translate natural language queries into the corresponding database query language. Recent methodologies, such as \cite{gao2023texttosql,pourreza2023dinsql,dong2023c3}, have attempted to leverage Large Language Models (LLMs) for the Text-to-SQL task. Nevertheless, the direct transfer of methods from Text-to-SQL to NL2GQL proves challenging due to differences in syntax structures, relationship representation, and database schema between GQL and SQL\cite{guo2022spcql}. In their most recent work, \citet{zhou2023r3nl2gql} utilize both smaller and larger Foundation Models as rerankers, rewriters, and refiners. To assess the effectiveness of their approach, they constructed a bilingual dataset. While our approach aligns with theirs in terms of the methodological flow, there are substantial disparities in the NL2GQL approach and dataset construction.

% \noindent \textbf{Fusing LLMs and Structured Databases.}
\subsection{Fusing LLMs and Databases}
% \yscomment{Discuss the existing studies trying to combine LLMs with graphs, databases. We can talk about RAG techniques, we can talk about existing prompting methods of KBQA. but we eventually highlight that we deeply fuse the LLMs and graph DB by proposing strategies for both fine-tuning and prompting.}

LLMs have shown its powerful capabilities in a diverse range of NLP tasks~\cite{min2023recent,kaddour2023challenges,ferrara2023should}.
Incorporating LLMs with databases is an important research question~\cite{pan2024unifying,luo2023chatkbqa,gao2023texttosql,pourreza2023dinsql,zhou2023r3nl2gql}, which benefits to improve the efficiency of structured data processing and ease the hallucination issues of LLMs.
%Large language models, with their formidable comprehension and generation capabilities, are leveraged to tackle a diverse range of NLP tasks \cite{min2023recent,kaddour2023challenges,ferrara2023should}. 
%Structured databases \cite{chang2004structured}, owing to their efficient management and utilization of data, play an increasingly vital role in information systems by facilitating the organization, storage, retrieval, and analysis of diverse data types. In recent years, the fusion of LLMs and structured databases has garnered growing attention and research \cite{pan2024unifying,luo2023chatkbqa,gao2023texttosql,pourreza2023dinsql,zhou2023r3nl2gql}. This integration brings about various benefits, such as improving the efficiency of structured data processing, enhancing the semantic comprehension abilities of LLMs, and reinforcing the interaction between NLP models and databases.
The integration of LLMs and structured DB is primarily approached through two main paradigms. 
The first paradigm extracts the information from databases to enhance the contexts for prompting LLMs in generation. 
For instance, techniques like Retrieval Augmented Generation (RAG) \cite{gao2023retrieval} can be employed to retrieve and query relevant data from databases, thereby enhancing the generated results \cite{kang2023knowledge, wang2023knowledgpt, modarressi2023ret, luo2023chatkbqa, yang2023llm, gu2023few, zhou2023r3nl2gql} during inference. 

The second paradigm injects the knowledge of databases into LLMs by fine-tuning~\cite{de-cao-etal-2021-editing,lin-etal-2021-knowledge,sen:nlrse2023,kang2023arxiv}.
% \yscomment{Cite papers like knowledge edit and knowledgpt}.
A key aspect of this approach is to make controllable edit to the parameters in LLMs.
Our approach take advantages of both paradigm.
We tuned a foundation LLM to align it to knowledge of a graph DB on the one hand.
On the other hand, we elaborately extract relevant schema from the graph DB for prompting, which deeply fuses LLMs and databases together.
%The second paradigm involves harnessing the powerful comprehension and generation capabilities of LLMs to aid in the more efficient utilization of databases. A key aspect of this approach is generating executable database query language based on the natural language and schema of databases \cite{}. Our approach also belongs to the second category. Initially, we utilize prompts to extract data from the database using ChatGPT. Subsequently, through fine-tuning LLMs, we ensure thorough alignment between LLMs and structured databases, resulting in a more accurate generation of database query language.

\section{Preliminary}

\subsection{Graph Databases}

A graph database $\mathcal{G}$ is a relational database, which is stored in the form of $\mathcal{G} =\{(s, r, o)| s, o \in \mathcal{E}, r \in \mathcal{V})\}$.
Here $\mathcal{E}$, $\mathcal{V}$ denotes the vertex set and the edge set, respectively. 
Graph database is a property graph, where every vertex and edge can be defined with \textit{labels} and have any number of \textit{properties} (aka \textit{attributes})~\cite{besta:cs2023}.
Each graph database is associated with a schema $\mathcal{S}$.

\subsection{NL2GQL with LLMs}

%Following existing study~\cite{}\yscomment{Cite papers}, we formally define the task of NL2GQL as follows.
%Given a graph database $\mathcal{G}$. 
When it comes a natural language query $Q$ inquiring about the information in $\mathcal{G}$, NL2GQL task requires a system to transfer $Q$ into a graph database language query $L$ (e.g., Cypher, nGQL, and Gremlin)
% \yscomment{We need to give a complete example of GQL in the introduction or somewhere.}, 
which can be executed over $\mathcal{G}$ to retrieve the predicted answer.
In this paper, we treat an LLM as an NL2GQL system to map $Q$ to $L$ with the consideration of pre-defined $\mathcal{G}$.
\section{Our Method}

%\yscomment{Overview}

In order to align LLMs to a Graph DB and achieve NL2GQL with high accuracy, we propose a pipeline as illustrated in Figure~\ref{fig:Overview}.
% \yscomment{Please redraw the figure by using the module names we renamed.}.
Initially, we \textit{generate NL-GQL pairs as instruction} for fine-tuning the LLM.
Specifically, starting from the Graph DB, we collect a set of canonical NL-GQL templates annotated with GQL.
Then we extend these seed pairs to a set of NL-GQL pairs with a wide query coverage and diverse expressions via two-step data augmentation based on ChatGPT.
These generated NL-GQL pairs are further grounded and verified.
Next, we \textit{fine-tune LLMs via LoRA} with the generated instruction such that the tuned LLM will be aligned with the knowledge in the graph DB.
For inference, to bound the LLMs with the graph DB schema, we \textit{organize the relevant schema to the query as the context of the prompt} and request the LLM to produce GQL using the identified schema.
% \krcomment{Maybe the expression here needs to correspond to the section name below and the figure.}

\subsection{Generating NL-GQL Pairs as Instructions}
\label{sec:nlgql_pair}

% \yscomment{Revise the Section 8 here.}
% \yscomment{Please rewrite this paragraph. 1. describe the challenge for generating NL-GQL pairs as instruction. 2. introduce how we solve the challenges and give an overview to this module which connects the roles of the subsection you would like to introduce. 3. Always simulate how people write (the token they use, the layout of their paragraph and their logics.) the similar procedure~\url{https://arxiv.org/pdf/2011.07743.pdf}, Section 3.1}

%Constructing NL-GQL data pairs from a graph dataset is an inherently challenging task. It requires expertise to accurately annotate each NL query with its corresponding GQL. Additionally, collecting large-scale human-annotated NL-GQL data pairs is time-consuming and labor-intensive. Moreover, ensuring the diversity and comprehensiveness of NL data, covering a wide range of query types, is paramount. Ultimately, the utmost importance lies in guaranteeing that the generated GQL queries are not only effective and precise but also seamlessly align with the intent behind the NL queries.
It is non-trivial to construct NL-GQL data pairs deriving from a given $\mathcal{G}$.
Different from the majority of existing QA over DB challenges like Spider~\cite{yu2018spider}, GrailQA~\cite{gu2021beyond} and SpCQL~\cite{guo2022spcql}, where the narrative questions are associated with a database in general domain.
$\mathcal{G}$ is in specific domain such that it is not applicable to employ the above datasets as initialization.
In addition, it is impossible to be scaled up if we solely rely on human labor to annotate the NL-GQL pairs in a specific domain.
Therefore, it is vital to develop a labor-saving pipeline to generate high-quality NL-GQL pairs that are able to cover a wide range of query types on a specific domain.

%LLMs are believed to surpass humans as annotators in various scenarios, thanks to their powerful understanding and generation capabilities \cite{gilardi2023chatgpt}. 
%Therefore, we have chosen to utilize LLMs to generate NL-GQL data pairs. 
% \xtycomment{This paragraph can be shortened. Many details are already in the following paragraphs.}
Therefore, we employ ChatGPT to play an important role in the pipeline of NL-GQL pairs generation.
With the awareness that ChatGPT have lack of the specific knowledge of $\mathcal{G}$, 
a \textit{NL-GQL template collector} module is developed to collect the canonical NL-GQL templates which decouple the redundant schema items from the GQL.
Then, a \textit{two-step instruction generator with self-instruct} module is implemented. 
Its first step strengthens the guidance to ChatGPT on $\mathcal{S}$ via providing the schema of graph DB as well as canonical NL-GQL templates as context for prompting.
In its second step, the outputs are verified via GQL2NL with reasoning the intermediate steps.
%We simplify the complexity of NL-GQL generation by explicating the intermediate steps.
Furthermore, we continue to improve the quality and coverage of instructions by self-supervising the instruction tuning process until the desired number of NL-GQL pairs is obtained.
Eventually, we ground the NL-GQL pairs via a \textit{graph DB grounding} module.
%In particular, a \textit{NL-GQL template collector} module collects the canonical NL-GQL templates as initial prompts to facilitate LLMs to expand data in size.
%Then, a \textit{two-step instruction generator with self-instruct} module is implemented. 
%The first step focuses on generating NL-GQL pairs with broader coverage.
%The second step aims to verify the correctness of the generated NL given the GQL, ensuring consistency between NL and GQL.
% NL with enriched expressions, 
% NL-GQL pairs generation on the breadth aims to generate more questions with enrich expressions.
% GQL2NL generation on the depth aims to generate more complicated questions with multi-hop reasoning
%\yscomment{Revise later.}.
%Eventually, we ground the NL-GQL pairs via a \textit{graph DB grounding} module.

% In parallel, we address the challenges of data diversity and comprehensiveness, as well as ensuring consistency between GQL and NL, by using the \textbf{Two-channel instruction generation with self-instruction}. 
% Additionally, \textbf{Graph DB grounding} is employed to ensure the effectiveness and precision of GQL.

\begin{spacing}{1.5}
\end{spacing}
\noindent \textbf{NL-GQL template collector.}
% \yscomment{To enhance the connection to the section title, change to ``Canonical NL-GQL template collection''}\yscomment{We need to highlight we collect templates instead of questions so we do not need to show any original NL-GQL below.}
% \yscomment{Remember do not directly say what we do, elaborately describe why we do so.}
% A body of research has demonstrated that in-context learning significantly enhances the remarkable performance of LLMs on downstream tasks \cite{brown2020language,liu2021makes,zhao2021calibrate}. Therefore, we have collected canonical NL-GQL templates to demonstrate the efficacy of in-context learning.
% Ensuring the representativeness of canonical NL-GQL templates is crucial for achieving diversity and comprehensiveness in the generated dataset. 
To ensure the diversity and comprehensiveness of generated NL-GQL on different graph DB, we formulate an NL-GQL template as a narrative question with masked schema items (i.e., entities and properties) followed by a GQL skeleton, where the entity labels and properties of the node are masked out.
An example is shown below:
\begin{itemize}[label={}, labelsep=0pt, leftmargin=15pt]
\item {\textit{What is [entity]'s [property]?
%\textcolor{blue}{\begin{CJK*}{UTF8}{gbsn}[entity]的[property]是什么？\end{CJK*} }
}}
\vspace{-5pt}
\item {
\textit{MATCH (s:node\{name:`[entity]'\})}

\vspace{-5pt}
\textit{RETURN s.node.[property]}
}.
\end{itemize}
where we employ the placeholder \texttt{[entity]} to denote the named entities and \texttt{[property]} to denote the property of an entity in both the NL and GQL. 

We define $8$ types of the canonical NL-GQL templates covering queries about \textit{entity property}, \textit{numerical sorting}, \textit{attribute comparison} and so on\footnote{The detailed query types and their corresponding templates can be found in Appendix \ref{query_type_show}}.
For each NL type, we manually annotate $K$ data pairs and include them into a data pool $\mathcal{D}$.
As we can see, the NL-GQL templates are able to decouple the schema-related information for the queries while keeping the main semantics and syntax, which can be adapted to different domains. 
%Initially, NL queries are categorized based on the graph data schema, with general NL types outlined in Table \ref{table:question_types}. 
%Adjustments can be made for different schema, as some schema may not contain as many NL types.
% \yscomment{where does the typical type come from? can we cite some papers indicating that we follow previous type definition.},\yscomment{what kind of adjustment?}. 
% \yscomment{Remove the original NL-GQL below.} 

%Notably, in the NL-GQL templates, named entities and properties are masked using special characters.
%For instance:
% \begin{itemize}[label={}, labelsep=0pt, leftmargin=15pt]
% \item {
% \begin{CJK*}{UTF8}{gbsn}国泰君安的股票代码是什么？\end{CJK*} 

% \vspace{-5pt}
% \small
% \textcolor{blue}{What is Guotai Junan's stock code?}
% }

% \item {
% \texttt{MATCH (s:stock{name:`GuotaiJunan'}) }

% \vspace{-5pt}
% \texttt{RETURN s.stock.code}
% }.
% \end{itemize}
% The masked NL-GQL data pair would appear as follows:

% \begin{itemize}[label={}, labelsep=0pt, leftmargin=15pt]
% \item {
% \begin{CJK*}{UTF8}{gbsn}[s]的股票代码是什么？\end{CJK*} 
% \vspace{-5pt}
% \small
% \textcolor{blue}{What is [s]'s stock code?}
% }
% \item {
% \texttt{MATCH (s:stock{name:'[s]'})}

% \vspace{-5pt}
% \texttt{RETURN s.stock.code}
% }.
% \end{itemize}

% \yscomment{Can we present the template in a more general way as follows?}

\begin{table}[t!]
\centering  
\tabcolsep=0em
\small
\begin{tabular}{l}
\toprule
%\multicolumn{1}{c}{\textbf{NL-GQL Pairs Generation Prompt}} \\ \midrule
\begin{tabular}{l}
%\begin{CJK*}{UTF8}{gbsn} 知识图谱的模式为：\end{CJK*}\{Graph Schema\}\\ 
\texttt{[Schema of Graph DB]}:  \\
\quad \textit{ Nodes}\\
 \quad \quad \textit{  \{'tag': 'stock', 'properties': [('code', 'string'), …\}}\\
 \quad \quad \quad ... \\
\quad \textit{ Edges}\\
  \quad \quad \textit{ \{'edge': 'has\_stock\_data','start\_tag':'stock', }\\
  \quad \quad \qquad \qquad \textit{ 'end\_tag':'stock\_data','properties': []\}}\\
 \quad \quad \quad ... \vspace{5pt}\\
%\begin{CJK*}{UTF8}{gbsn}任务: \{任务的描述\end{CJK*}\}\\ 
\texttt{[Task Description]: } \textit{Please generate a new NL-GQL}\\
\textit{template data pair of the same type as the provided NL} \\
\textit{based on the given Graph database schema and NL-GQL} \\
\textit{demonstrations. In the ...}\vspace{5pt}\\
\texttt{[NL-GQL]: } \\
\quad \textit{What is \texttt{[entity]}’s stock code?} \\
%\begin{CJK*}{UTF8}{gbsn}Examples：\end{CJK*}\\
%NL:{[}s{]} \begin{CJK*}{UTF8}{gbsn}的股票代码是什么？\end{CJK*}\\ 
%\textcolor{blue}{\texttt{(What is [s]’s stock code? )}}
\quad \textit{MATCH (s:stock\{name:\texttt{'[entity]'}\})} \\  
\quad \textit{RETURN s.stock.code} \\                  
\quad ... \\

%\texttt{[Output]: } 
\end{tabular} \\
\bottomrule
\end{tabular}
\caption{The prompt for NL-GQL data pair generation.}
% \yscomment{give an example of task description}} 
% task description is too long in appendix \ref{task_desc}

%Notably, the blue-colored text in the prompt as the English translation of the Chinese and is not an inherent component of the prompt.
% \yscomment{Please give some examples to show the organization of schema. Take query type as a variable in the prompt just like [s] to write.}\yscomment{where are the demonstrations and what is the last token we use to indicate LLMs to generate?}\yscomment{We do not have WHERE function when generating NL-GQL pairs?}}
\label{table:data_pair_generate}
\vspace{-1.5em}
\end{table}

\begin{spacing}{1.5}
\end{spacing}
\noindent \textbf{Two-step instruction generator with self-instruct.}
% \yscomment{Change to ``Two-channel instruction generation with self-instuct''}
% \yscomment{Always describe why you employ/propose such a method. Why two-channel?}
%Self-instruct \cite{wang2023selfinstruct} has confirmed the efficacy of generating high-quality and diverse data from seed data using LLMs. Thus, we utilize a two-channel instruction generation approach with self-instruct to generate a broad range of query coverage and diverse expressions in NL-GQL data pairs.
% \yscomment{What is the goal of this channel? Based on this goal why we have to choose self-instruct?}. 
The first step aims to generate candidate NL-GQL pairs with $\mathcal{S}$ as well as $\mathcal{D}$ and using ChatGPT.
We design a unified prompt but adjust to the different types of query by changing instructions for each category.
As shown in Table~\ref{table:data_pair_generate}, each prompt consists of the following components:
\begin{itemize}[leftmargin=*]
		\setlength{\itemsep}{0pt}
\item \texttt{[Schema of Graph DB]}: According to \citet{gao2023texttosql}, including the DB schema in the prompt helps the generation of SQL that fits the DB. 
Therefore, our prompt used for generating NL-GQL data pairs consists of the entire schema of the graph DB. 
Specifically, we organize the schema by flattening the nodes set into a sequence, wherein each node type is represented as a key accompanied by its corresponding properties. Similarly, each edge is defined with its start and end node types, along with any associated properties, as depicted in Table \ref{table:data_pair_generate}.

% \yscomment{Describe how we organize the schema.}
 \item \texttt{[Task Description]}: It is a textual sequence describing the query type of pairs to be generated, we pre-define the description in Appendix~\ref{task_desc}.
 % \yscomment{Is it a pre-defined sequence of text or something? please describe it.}
\item \texttt{[NL-GQL]}: We sample $K$ examples of a certain type from $\mathcal{D}$ and randomly sample properties from the schema to instantiate the \texttt{[property]} variable, resulting in different demonstrations.
%\item \texttt{[Output]}: This is the final token we use to signal LLMs to begin generating.
% \yscomment{Do we fill in the labels or properties as described?}.
\end{itemize}
%We generate questions according to the type of NL one by one. 
%For each query type, $K$ examples of the certain type are randomly selected from $\mathcal{D}$.
%, where the type of NL in each example is the same as the type being generated.
 % \yscomment{You mentioned K data pairs so how can you randomly select K examples for each question?}. 
% \yscomment{But in template, we already decoupled the schema items from the pairs, what kind of schema we used for pre-pending the prompt? Here, we use the entire graph schema or partial of them?}, 
%a task description,
% \yscomment{What kind of task is? Do we have different task description in this channel?}, 
%and demonstrations.
% \yscomment{The demonstration we used here is templates with randomly inserted schema items? Please mention it or the readers do not know what it is.}.
% By utilizing the entire graph schema, our objective is to generate data that encompasses a wide range of scenarios and comprehensively represents the information within the schema.
% The task description describes the data to be generated, thereby defining the types of NL-GQL template data to be generated.
% In the demonstrations, NL-GQL template data is utilized, with all NL templates belonging to the same type.
 % After prompting LLMs, M NL-GQL pairs are generated\footnote{The detailed prompt design for NL-GQL data pair generation is presented in the Appendix. ~\ref{}.}. 
In this case, we can prompt LLMs to generate NL-GQL pairs that are related to the graph DB.
For simplicity, we denote this procedure as:
 \begin{align*}
  Q', L' = \text{NL-GQL-Gen}(\mathcal{S}, \mathcal{D}).
 \end{align*}
%The specific format for building the prompt is depicted in Table~\ref{table:data_pair_generate}. The self-instruct method we utilize is expressed as $Q', L' = \text{Self-Instruct}(\mathcal{S},Q, L)$, where Q represents the NL query, L represents the corresponding GQL, and $\mathcal{S}$ represents the schema of the graph database. The generated $Q'$ will be populated with actual entity names and properties, executed, and correct $Q'$ and $L'$ pairs will be added to the pool. 
%In the next step, the output $L'$ will serve as the input.

% \yscomment{Do we take the generated data pairs as the demonstration for next round og generation? How to ensure the validation of $Q'$ and $L'$ generated at this stage?}
 % \yscomment{Instead of describing purely using text, you can involve the math symbols to make everything clearer. For example, $Q', L' = \text{Self-Instruct}(Q, L, \mathcal{S})$ if the input includes $Q, L, \mathcal{S}$.}\yscomment{Also, you did not mention about how the $\mathcal{S}$ and $\mathcal{G}$ influence the procedure, we did not consider them in this stage at all?}

% \yscomment{Where we get the GQL for this step?}. 

The second step takes $L'$ as the input and aims to verify NL with GQL generated from the first step.
Several studies \cite{wei2022chain, kojima2022large, zhou2022least, diao2023active} have established the efficacy of utilizing CoT \cite{NEURIPS2022_9d560961} to yield more accurate answers by generating intermediate reasoning steps. 
In this step, we apply CoT to conduct GQL-to-NL, the goal of which lies in verifying the quality of the generated NL given a GQL.
%GQL-to-NL represents a form of inferential generation task, deducing NL that is able to align with a given GQL.
%In this channel, we construct a prompt consisting of schema, instructions, and CoT demonstrations.
The prompt format closely resembles that of Table \ref{table:data_pair_generate}, with the task-adjusted description and replacement of \texttt{[NL-GQL]} with \texttt{[CoT-based GQL2NL]} demonstrations:
\begin{itemize}[leftmargin=*]
\item \texttt{[CoT-based GQL2NL]}: We take the $K$ examples in the first step but present the transition from GQL to NL in the format of a explicit reasoning chain.
Specifically, we break down the GQL into individual keywords function clauses based on keywords such as \textit{MATCH}, \textit{WHERE} and \textit{RETURN}.
For example, the given GQL is \textit{MATCH (s: stock\{name: `\texttt{[entity]}'\}) RETURN s.stock.code}. Its CoT process is as follows:
\textit{MATCH (s: stock\{name: `\texttt{[entity]}'\} identifies the entity with the stock name \texttt{[entity]}. RETURN s.stock.code returns the code of the stock.}
Combining these two parts, so the output is: \textit{"What is [entity]'s code?"}    
% \yscomment{give an example.}
% \yscomment{We do not have WHERE in the demonstration for the first step, how can we have WHERE function in $L'$?}.
% In table 1, we can see some GQLs  have WHERE ,
\end{itemize}
%with alterations confined to the task description and demonstrations.
%The demonstrations encompass GQL and the CoT reasoning process from GQL to NL.
% The detailed design of the GQL-to-NL prompt is presented in Appendix ~\ref{}
% Then, we identify the intentions of each clause before finally concatenating them into a complete NL query.
% The self-instruct method we utilize is expressed as $Q' = \text{Self-Instruct}(\mathcal{S},L)$, where L represents the corresponding GQL, and $\mathcal{S}$ represents the schema of the graph database. We will also iterate this method until a corresponding $Q'$ is generated for each $L$.
Therefore, we can obtain an alternative question by prompting LLMs with the given GQL.
For simplicity, we denote this procedure as:
 \begin{align*}
  Q^{''} = \text{GQL2NL-Gen}(\mathcal{S}, L').
 \end{align*}
To ensure the consistency of $L$ and $Q$ across all data entries,
%, we separately embed the $Q$ generated by the first and second channels for the same $L$. 
% \yscomment{How can we compare their similarity?}, 
we compare encode them via the sequence matching encoder and compare their similarities.
If it falls below a certain threshold, we consider the $Q'$ and $L'$ to be inconsistent, leading to the discarding of the respective data entry. 
%This threshold is established at $0.8$.
Otherwise, we include the NL-GQL pairs into data pool $Q', L' \in \mathcal{D}$.
%\yscomment{Based on our description, $L'$ excludes MATCH command, do we conduct further process to include the command?}.
% \yscomment{We describe self-instruct here.}

%We sequentially execute the first step and the second step, iterating through this process using self-instruct. Any inconsistencies between NL and GQL, GQL syntax errors, or repeated NL-GQL pairs are filtered out. High-quality NL-GQL pairs are then placed into the example pool. This iterative process continues until we generate the desired number of NL-GQL pairs.

The two step generation results in high-quality NL-GQL pairs.
To enhance the diversity of NL-GQL pairs, we conduct self-instruct~\cite{wang2022self} for the generation procedure, where the demonstrations used for the next iteration are extracted from $\mathcal{D}$. 
This iterative process continues until we generate enough NL-GQL pairs.

\begin{spacing}{1.5}
\end{spacing}
\noindent \textbf{Graph DB grounding.} 
In the previous steps, all generated NL and GQL pairs in $\mathcal{D}$ are templates with \texttt{[entity]} placeholders. 
Then, we replace these placeholders with named entities from the graph DB to obtain instantiation of NL as well as GQL.
The GQLs are executed over the associated graph DB to check if they are semantically and syntactically correct to retrieve the answers.
Only the executable NL-GQL pairs will saved.
As a result, we have a high-quality data pool in specific domain with each entry containing an NL, GQL, template NL, template GQL, and the corresponding executed answers. 
%In the domains of financial stocks and medical consultations, we've developed two specific datasets: FinGQL and MediGQL.
%We divided the dataset into train, dev, and test sets. The statistics of the datasets are in Table~\ref{dataset_static}.

\subsection{Aligning LLMs to a Graph DB with LoRA}

% \yscomment{Revise Section 9.3 here, one parageraph is enough.}

After we generate a set of NL-GQL pairs, we can fine-tune an existing foundation LLM to align with the graph DB.
%To align the LLM with the Graph DB, it is essential to inject the knowledge from the dataset constructed from the Graph DB into the model. We need to utilize this dataset for training the LLM.
%LLMs, with their extensive parameters, excel in various NLP tasks. However, their size poses challenges for customization, especially in resource-limited environments. 
To avoid extensive parameter tuning, we apply Parameter Efficient Fine-Tuning (PEFT) techniques, which is able to minimize fine-tuning parameters and optimize memory usage while maintaining the performance~\cite{liu2022few,hu2021lora}. 
%Among numerous PEFT techniques, LORA \cite{hu2021lora} undoubtedly stands out as a widely adopted and overall efficient method. 
We choose LoRA~\cite{hu2021lora} for fine-tuning the selected LLMs due to its wide application. 
Specifically, we feed the NL-GQL pairs obtained from Section~\ref{sec:nlgql_pair} into the LLMs.
For each pair, we extract the schema items in the GQL as the context to the NL as an instruction,
then supervised fine-tuning calculates the cross-entropy loss over the ground truth GQL.
In this way, the fine-tuned LLMs learn the schema of the domain-specific graph DB and understand the transition from the NL to the GQL.

%It's worth noting that during the fine-tuning stage, the relevant schema is extracted from the labeled GQL in the training data using regular expressions.
% \krcomment{Maybe the following should be included in Implementation Details section.}

\subsection{Organizing Schema for Prompting}
\label{related_schema_extraction}

% \yscomment{Revise Section 9.2 here}

% \krcomment{Maybe we should indicate the difference between training and inference since the following methods are for inference only.}
% 1、从知识图谱中抽取每个实体类型对应的命名实体并存为命名实体字典{"entity1":[name1,name2...],"entity2":[name1,name2...]...}
% 2、把抽取出来的所有的命名实体都加载到分词工具的字典里
% 3、使用分词工具对新question进行分词，并提取出包含的命名实体
% 4、用分词出来的命名实体去查找从命名实体字典中找到对应的实体类型
% 5、对抽取的实体类型进行过滤，因为同一个命名实体可能对应多个实体类型e1,e2(比如：在股票数据集中，张三既可以是基金经理又可以是董事长),把新question mask，得到masked question，然后和训练集里面的masked question进行相似度比对，得到最相似的masked question对应的实体类型e_lst。最后选择在e_lst中的实体类型，如果不在的话则把多个实体类型都保留。
% 6、对抽取出来的实体进行链路补全，添加涉及的relation和中间实体类型，从知识图谱的schema中抽取实体类型和relation对应的schema。

Regarding inference for NL2GQL tasks, we prompt the tuned LLM to generate $L$ given $Q$ directly.
Similar attempts have been carried on Text2SQL by representing a DB schema with a linear ``code'' sequence and appending the sequence as the context for prompting~\cite{nan2023enhancing}, we prepend the schema of the graph DB as context but differ in the way of formulation.
Specifically, we extract the key schema items that are relevant to the NL rather than the entire schema as the context.
To this end, a \textit{label linker}
% \yscomment{Change ``entity'' to ``label'' because it is a common concept in graph DB.} 
is used to extract the named entities from $Q$ and link them to the labels in the graph DB.
Then, a \textit{LLM prompt} organizes the extracted labels as the context and generates the main body of the GQL.
Eventually, a \textit{join-table linker} further infers the joint tables based on the extracted labels and append it to the main body to form a prompt.

%To verify that the relevant schema of the NL as context can enhance the NL2GQL model's capability, we need to extract the Relevant Schema based on NL. Firstly, we employ Entity Information Extraction from the graph database to extract information relevant to named entities. Subsequently, we utilize a Question Tokenization and Entity Type Inference process to infer the entity type information corresponding to the named entities in the NL. Finally, we conclude with Link Completion and Schema Retrieval to extract relevant schema information.

\begin{spacing}{1.5}
\end{spacing}
\noindent \textbf{Label linker}. 
% To accurately extract named entities in NL that are relevant to the graph DB, we require the tokenizer to recognize the named entities in the graph database. In this step, we extract named entities corresponding to each entity type from the graph dataset and store them in a dictionary. Subsequently, these named entities are loaded into the tokenization tool's dictionary in preparation for further processing.
To accurately identify named entities in NL and link them to the labels in the graph DB, we first collect all the labels from the graph DB and store them in a dictionary, where the labels are saved as the key and their corresponding properties are saved as the value.
%the key represents the schema entity type, while the value list comprises the properties of the schema entity and the named entities extracted from the graph dataset.
%\yscomment{If it is a dictionary, what is the key and what is the value?}
When there comes a $Q$, we tokenize the sentence and employ an off-the-shelf named entity tagger to identify the named entities within it.
Then, we map these named entities to possible keys in the dictionary via string match.
%\yscomment{how do we match them?}. We sequentially search through the value lists associated with each label key in the dictionary. If the named entity is found within any of these value lists, we record the corresponding label key.
This step may result in multiple ambiguous labels. 
For example, the named entity ``\textit{Sam}'' could be linked to either a fund manager or a chairman.
%We disambiguate the query by comparing similarities and extracting the NL-GQL data pair from the training set, with NL closest to the input NL. 
We disambiguate the named entities by choosing the labels in a GQL from $\mathcal{D}$, the NL of which has the highest similarity to $Q$.
%We employ regular expressions to extract the list of labels corresponding to the GQL. If any label from the multiple ambiguous labels is present in this label list, we choose that label. Otherwise, we retain all possible labels.

\begin{spacing}{1.5}
\end{spacing}
\noindent \textbf{Join-table linker}. 
Since many questions inquire about information that requires reasoning across multiple tables, but not all table information is explicitly mentioned in the question, it is necessary to perform table linking with the linked labels. 
We formulate the problem as searching for the shortest path through the tables where the identified labels are located.
% ~\yscomment{Is there any algorithm we applied to solve the problem like Dijkstra algorithm to the shortest path problem?}.
We utilize the A* algorithm \cite{hart1968formal} to address the shortest path problem. This results in the shortest chain linking all the identified labels, which serves as the MATCH keyword function in a GQL.
%Therefore, based on the extracted labels, we check if the chain of links is complete, adding intermediate labels for incomplete chains, and establishing linking relationships between labels. We will ultimately obtain the schema related to input NL. It's worth noting that since our dataset is in Chinese, we have provided corresponding Chinese annotations for all the fields in the tables.
%\yscomment{Revise later}.

\begin{table}[t!]
\centering  
\tabcolsep=0em
\small
\begin{tabular}{l}
\toprule
%\multicolumn{1}{c}{\textbf{NL-GQL Pairs Generation Prompt}} \\ \midrule
\begin{tabular}{l}
%\begin{CJK*}{UTF8}{gbsn} 知识图谱的模式为：\end{CJK*}\{Graph Schema\}\\ 
\texttt{[Relevant Schema of Graph DB]}:  \\
\quad \textit{ Nodes}\\
 \qquad \quad \textit{  \{'tag': 'stock', 'properties': [('code', 'string'), …\}}\\
 \qquad \quad \quad ... \\
\quad \textit{ Edges}\\
  \qquad \quad \textit{ \{'edge': 'has\_stock\_data','start\_tag':'stock', }\\
  \qquad \quad \qquad \qquad \textit{ 'end\_tag':'stock\_data','properties': []\}}\\
 \qquad \quad \quad ... \vspace{5pt}\\
%\begin{CJK*}{UTF8}{gbsn}任务: \{任务的描述\end{CJK*}\}\\ 
\texttt{[Task Description]: } \textit{You are a graph database expert. }\\
\textit{Please write the corresponding graph query language based }\\
\textit{on the relevant schema and natural language.}\vspace{5pt}\\
\texttt{[NL]: } \textit{Which stocks have a minimum price below 20 and}\\
\textit{ an opening price below 30 today?} \\
%\texttt{[Output]: } 
\end{tabular} \\
\bottomrule
\end{tabular}
\caption{The prompt for fine-tuning LLMs.}
\label{tab:prompt_peft_llm}
\vspace{-1.5em}
\end{table}
The MATCH keyword function of inferred GQL is verified by the outputs of joint-table linker to form an executable GQL.
% \yscomment{My understanding is that join-table linker goes after LLM prompt as LLM prompt generates WHERE, RETURN functions, and we need to concatenate them with the MATCH function to form a complete GQL.}
% 我们在造数据部分的最后一步就是使用真实的命名实体的属性回填templete了，在LLM的微调和推理部分都是使用的完整的NL-GQL pair. 链路补全只是在抽取relevant schema作为NL的context引导LLM输出时才使用。

\begin{spacing}{1.5}
\end{spacing}
\noindent \textbf{LLM prompt.} 
% We organize the extracted labels in the prompt as follows
After acquiring the relevant schema, incorporate the task description and input natural language to compose a comprehensive prompt, as shown in Table \ref{tab:prompt_peft_llm}.

\section{Experiments}

\subsection{Experimental Setup}

\begin{table}[t!]

\centering
\tabcolsep=0.25em
\small
\begin{tabular}{l ccc ccc}
\toprule
\multicolumn{1}{l}{\multirow{2}{*}{\textbf{Datasets}}} & \multicolumn{3}{c}{\textbf{FinGQL}} & \multicolumn{3}{c}{\textbf{MediGQL}} \\
\cmidrule(lr){2-4} \cmidrule(lr){5-7}
                      & \textbf{train}    & \textbf{dev}     & \textbf{test}   & \textbf{train}    & \textbf{dev}    & \textbf{test}    \\ 
                     \midrule
                     
\multicolumn{7}{c}{\textit{Graph DB info}}\\ \midrule
\#Nodes & \multicolumn{3}{c}{$668508/12$ types} & \multicolumn{3}{c}{$44656/8$ types} \\
\#Edges & \multicolumn{3}{c}{$730583/14$ types} & \multicolumn{3}{c}{$312159/12$ types} \\
\#Properties & \multicolumn{3}{c}{$62$ types} & \multicolumn{3}{c}{$14$ types} \\ 

\midrule

\multicolumn{7}{c}{\textit{Dataset info}}\\ \midrule

% \#Nodes& $12658$ & $1780$ & $3622$ & $10548$ & $1564$ & $2937$\\
% \#Edge& $5683$ & $781$ & $1627$ & $5170$ & $768$ & $1404$\\
\#Nodes (Avg.)& $1.81$ & $1.78$ & $1.81$ & $1.76$ & $1.84$ & $1.73$\\
\#Edge (Avg.)& $0.81$ & $0.78$ & $0.81$ & $0.86$ & $0.90$ & $0.83$\\
\#Unique Template & $1434$ & $229$ & $419$ & $1907$ & $293$ & $590$\\
\#NL-GQL  & $7000$     & $1000$    & $2000$   & $6000$     & $850$    & $1700$    \\ \bottomrule
\end{tabular}
\caption{\label{dataset_static}
Statistics of the generated datasets.
% \yscomment{Can we add more statistics into the table to show the information of the graph DB.}
}
\vspace{-1.5em}
\end{table}

\noindent \textbf{Evaluation Set Construction.}
To verify the way we fine-tune a foundation LLM and prompt the tuned LLM for solving NL2GQL over a domain-specific graph DB, we choose StockKG\footnote{We collect financial data from open stock exchange websites: \url{http://www.sse.com.cn/market/price/report/}, \url{https://www.szse.cn/market/overview/index.html}, \url{https://sc.hkex.com.hk/TuniS/www.hkex.com.hk/Market-Data/Securities-Prices/Equities}} and DiseaseKG\footnote{\url{http://openkg.cn/dataset/disease-information}} as the graph DB in finance and medicine domains, respectively.
We follow the pipeline description in Section~\ref{sec:nlgql_pair} to generate a number of NL-GQL pairs.
% We randomly partition the data pool into training, development and test with ratio $xxx$\yscomment{The ratio we sample to form the test set}.
We randomly partitioned the data pool into training, development, and test sets with ratios of $7:1:2$, respectively.
The correctness of test sets are further verified manually.
These two data sets, namely \textbf{FinGQL} and \textbf{MediGQL}, have statistics shown in Table~\ref{dataset_static} and can be leveraged to support future research on NL2GQL over domain-specific graph DB\footnote{More details about the presented datasets are provided in Appendix \ref{dataset_detail_show}.}. We will release the datasets upon acceptance.
%We manually verify the correctness of test set, which results in NLs with annotated GQL.

% \yscomment{what $\mathcal{G}$ we used to start our experiment and how we collect the evaluation set.}

\begin{table*}[ht!]
\centering
\small
\tabcolsep=1.5em
\begin{tabular}{l l  c c  c c }
\toprule
\multirow{2}{*}{\textbf{Method}} & \multirow{2}{*}{\textbf{Backbones}} & \multicolumn{2}{c}{\textbf{FinGQL}} & \multicolumn{2}{c}{\textbf{MediGQL}}\\
 \cmidrule{3-4} \cmidrule{5-6}
& & \textbf{EM} & \textbf{EX} & \textbf{EM} & \textbf{EX} \\
\midrule

%  & Method  & \multicolumn{2}{c}{FinCQL} & \multicolumn{2}{c}{MediCQL} \\
% \midrule
%  & & EM & EX & EM & EX \\
% \midrule
\multirow{2}{*}{Fine-Tuning}  & mbart-large  & 69.55 & 53.45 &64.92  & 54.56    \\
                              & mt5-large        & 73.75 & 67.75 & 79.69 &  65.04  \\ 
\midrule
\multirow{5}{*}{In-Context Learning} 
                              & Baichuan2-13B-chat       & 18.20 & 14.90  & 23.54  &18.24  \\
                              & Chatglm3-6B        &  16.10 & 17.75 &18.78 &16.13   \\
                              & Qwen-14B-Chat         &  21.10  &22.75 &  24.24 & 21.07 \\
                              & ChatGPT-3.5    &40.65  & 40.10  &   49.38 & 41.38 \\
                            
\midrule
\multirow{3}{*}{Ours } 
         & Baichuan2-13B-Chat  & 78.10 & 71.25& 84.54  & 71.38  \\
        & Chatglm3-6B         & 77.30   &  68.95 & 83.81 & 70.51\\ 
        & Qwen-14B-Chat         & \textbf{79.65}    & \textbf{73.75}  &\textbf{86.05} &\textbf{72.13} \\ 
\bottomrule
\end{tabular}
\caption{\label{main_result}
Comparison between our method and the baseline method. The bold numbers indicate the best results.
 %\yscomment{Do we have more baselines the results of which are lower than us?}
}
\vspace{-1.5em}
\end{table*}

\begin{spacing}{1.5}
\end{spacing}
\noindent \textbf{Evaluation Metrics.}
We follow the evaluation methods of text2SQL, which include exact-set-match accuracy (\textbf{EM}) and execution accuracy (\textbf{EX})~\cite{yu2018spider}. EM measures the consistency of the individual components split by keywords within the predicted query and its corresponding ground truth, whereas EX compares whether the execution results on the database are consistent between the two.

\begin{spacing}{1.5}
\end{spacing}
\noindent \textbf{Baseline Methods.}
To validate and compare the effectiveness of our method, we selected two categories of methods as baselines:
\begin{itemize}[leftmargin=*]
\setlength{\itemsep}{0pt}
    \item \textit{Fine-tuning methods}.
    Instead of constructing instructions for fine-tuning, we employ the traditional approach~\cite{lu-etal-2022-unified,guo2022spcql} to fine-tune a pre-trained model, where the NL from NL-GQL data pairs serves as input, and the GQL serves as the label for training.
    % \yscomment{How do we finetune the models, please specify the distinction to our method.}
    \item \textit{In-Context Learning methods}. We follow the RAG paradigm to include the schema of graph DB as the context and sample $K$ examples with large semantic similarities to the NL as the demonstrations for prompting LLMs during inference \cite{dong2023survey,gao2024retrievalaugmented}.
\end{itemize}
%Firstly, we employed \textit{fine-tuning methods}. Since the datasets are in Chinese, we opted to use two widely used pre-trained models, mbart-large and mt5-large, as backbone models.
%Secondly, we explored the \textit{ICL methods}. We were interested in observing the ability of ICL in solving NL2GQL tasks. When selecting examples, we followed the approach outlined in \cite{liu2021makes}, choosing K examples most similar in embedding to a given question to construct demonstrations.

\begin{spacing}{1.5}
\end{spacing}
\noindent \textbf{Implementation Details.}
In selecting the backbone LLMs, we prioritize LLMs compatible with both Chinese and English due to the language of employed Graph DB.
As a result, we choose \texttt{Baichuan2-13B-Chat}~\cite{yang2023baichuan}, \texttt{Chatglm3-6B}~\cite{zeng2022glm}, and \texttt{Qwen-14B-Chat}~\cite{bai2023qwen} as the foundation LLMs to be aligned.
Regarding fine-tuning methods, we select \texttt{mbart-large}~\cite{tang2020multilingual} and \texttt{mt5-large}~\cite{xue-etal-2021-mt5} as the pre-trained models due to their wide application in multilingual scenarios.
Regarding the sequence matching for demonstration selection, we utilized the \texttt{all-MiniLM-L6-v2} checkpoint from the Sentence-Transformers library to encode a sequence. 
The threshold for verifying the NLs from two steps in NL-GQL pair generation is set as $0.8$.
All the number of demonstrations $K$ are set as $8$. 
% In Section \ref{sec:nlgql_pair}, the similarity comparison threshold is established at $0.8$\yscomment{Why we need a threshold here?}.
%In selecting the backbone model, we prioritize models compatible with Chinese and possessing relatively small parameter sizes. Options meeting these criteria include Baichuan2-13B-Chat \cite{yang2023baichuan}, Chatglm3-6B \cite{zeng2022glm}, and Qwen-14B-Chat\cite{bai2023qwen}. 
Our experiments were implemented on an A800 GPU.

% \begin{table*}[ht!]
% \centering
% \small
% \begin{tabular}{l|cc|cc|cc|cl}
% \toprule
% Query type & \multicolumn{2}{c|}{Entity property }        & \multicolumn{2}{c|}{Numerical sorting }    & \multicolumn{2}{c|}{Relationship inference } & \multicolumn{2}{c}{Yes/No question } \\
% Metrics    & EM                       & EX                     & EM                      & EX                    & EM                      & EX                      & EM                  & EX                  \\
%            &   81.87 &80.63     & 84.88  & 82.27   &  84.18   & 84.48                        &   69.40    &   68.87                  \\ \bottomrule
%            \toprule
% Query type & \multicolumn{2}{c|}{Relationship filtering } & \multicolumn{2}{c|}{Attribute comparison } & \multicolumn{2}{c|}{Edge property }          & \multicolumn{2}{c}{String filtering } \\
% Metrics    & EM                       & EX                     & EM                      & EX                    & EM                      & EX                      & EM                  & EX                  \\
%            &  79.55                        &    77.27                    &  76.01                       &     78.50                  &    70.15                     &       71.61                  &     88.89                &     85.19                \\ 
%           \bottomrule
% \end{tabular}
% \caption{\label{breakdown_result}
% Performance of our method on various types of queries in the FinGQL dataset.}
% \end{table*}

\subsection{Main Results}
Table \ref{main_result} illustrates a performance comparison between our method and the baseline methods. From the experimental results, we can draw the following conclusions:

a) Our approach demonstrates a significant improvement over the best baseline methods. 
Specifically, on the FinCQL dataset, it outperforms them by 5.90\% on the EM metric and by 6.00\% on the EX metric. Similarly, on the MediCQL dataset, it surpasses them by 6.36\% on the EM metric and by 7.09\% on the EX metric.

b) Backbone model is crucial for the NL2GQL task.  
The mt5-large model consistently outperforms the mbart-large model. Additionally, both the ICL method and our approach demonstrate that the Qwen-14B-chat model yields significantly better results compared to other open-source LLMs. Furthermore, within the ICL method, ChatGPT-3.5 exhibits notably superior performance compared to several other LLMs, although it only achieves roughly half the accuracy of our approach.

c) It's evident that a simplistic ICL method isn't suitable for NL2GQL. This may be attributed to the lack of domain knowledge from the data within the graph DB. To improve the effectiveness of the ICL method, further research is needed in prompt design to better guide LLMs in extracting domain knowledge.

Furthermore, it is evident from the result table that EM values generally exceed those of EX. EM serves as a more forgiving evaluation metric, while EX provides a more precise assessment by directly comparing GQL execution outcomes.

\subsection{Comparison with Domain-Specific LLMs}

% One of the reasons why the ICL performs poorly on the NL2GQL task is its insufficient understanding of the domain-specific knowledge. Therefore, we are interested in exploring whether Domain-Specific LLMs can perform better. The results in Table \ref{domain_model_compare} indicate that compared to the general-domain Chatglm3-6B and Qwen-14B-Chat, the domain-specific HuatuoGPT2-7B and BianQue-2 LLMs perform very poorly. It's possible that they focused too much on domain-specific knowledge and tasks, neglecting general knowledge and tasks.

One of the reasons why the ICL performs poorly on the NL2GQL task is its insufficient understanding of the domain-specific knowledge. Therefore, we are interested in exploring whether Domain-Specific LLMs can perform better. We experiment on two well-known medical-domain LLMs, HuatuoGPT2-7B~\cite{chen2023huatuogptii} and BianQue-2~\cite{chen2023bianque}, and the results are in Table \ref{domain_model_compare}. As shown in the table, compared to the general-domain Chatglm3-6B and Qwen-14B-Chat, the domain-specific HuatuoGPT2-7B and BianQue-2 LLMs perform very poorly. It is presumably because they focused too much on domain-specific knowledge and tasks, neglecting general knowledge and tasks. During the domain-specific supervised fine-tuning, the ability of language conversion of LLMs is largely compromised, resulting in poor performance in the NL2GQL task.

% \begin{table}[]
% \centering
% \small
% % \resizebox{\linewidth}{!}{
% \begin{tabular}{l c c}
% \toprule
% \textbf{Model}     & \textbf{EM} & \textbf{EX} \\ \midrule
% HuatuoGPT2-7B~\cite{chen2023huatuogptii}      & 0.059   &  0.059  \\
% BianQue-2~\cite{chen2023bianque}   &  0.18  &  0.18  \\ \hline
% Chatglm3-6B~\cite{zeng2022glm} &   16.10     & 17.75 \\
% Qwen-14B-Chat~\cite{bai2023qwen} &  21.10   &  22.75  \\ 
% \bottomrule
% \end{tabular}
% % }
% \caption{\label{domain_model_compare}
% The comparison between our method and two medical domain LLMs on the MediGQL dataset.}
% \end{table}

\begin{table}[]
\centering
\small
% \resizebox{\linewidth}{!}{
% \begin{tabular}{l l c c}
\begin{tabular*}{\hsize}{@{}@{\extracolsep{\fill}}l l c c@{}}
\toprule
\textbf{Domain}     & \textbf{Model}     & \textbf{EM} & \textbf{EX} \\ \midrule
\multirow{2}{*}{Medical domain} & HuatuoGPT2-7B      & 0.059   &  0.059  \\
~ & BianQue-2   &  0.18  &  0.18  \\ \midrule
\multirow{2}{*}{General domain} & Chatglm3-6B &   16.10     & 17.75 \\
~ & Qwen-14B-Chat &  21.10   &  22.75  \\ 
\bottomrule
\end{tabular*}
% }
\caption{\label{domain_model_compare}
The comparison of performance between medical domain LLMs and general domain LLMs using the ICL method on the MediGQL dataset.}
\vspace{-1em}
\end{table}

\subsection{Further Analysis}

\noindent \textbf{Efficiency Analysis.}
% To validate the inference speed of our method, we selected three schema-related approaches: one without schema, one with the complete Chinese schema, and one with the relevant Chinese schema. 
% Our experiments were conducted on an A800 GPU, performing batch predictions with a batch size set to 8 on FinCQL dataset. We randomly sampled 1000 test cases and computed the average inference time for each approach.
To observe the inference speed of our method, we conducted experiments on the test set of FinGQL and recorded its execution time. 
The results are shown in Table \ref{exe_time}. Based on Table \ref{exe_time} and Table \ref{finetuned_method}, it is evident that our strategy of incorporating the relevant Chinese schema leads to a substantial enhancement in effectiveness, accompanied by a slight increase in inference time. However, adding the full schema will significantly increase the inference time, and including unrelated distracting information may also lead to a decrease in accuracy.

\begin{table}[]
\centering
\small
% \begin{tabular}{l c}
\begin{tabular*}{\hsize}{@{}@{\extracolsep{\fill}}l c@{}}
\toprule
\textbf{Method}                    & \textbf{Execute time (Sec.)} \\
\midrule
Without Schema          &     0.77        \\
Full Schema             &        1.32      \\
Full Chinese Schema          &   1.52            \\
Relevant Schema          &       0.87    \\ 
Relevant Chinese Schema (Ours) &        0.88       \\ 
 \bottomrule
\end{tabular*}
\caption{\label{exe_time}
 Comparison of average inference times for various schema-related methods on the FinGQL dataset, measured in seconds.}
 \vspace{-1.5em}
\end{table}

\begin{spacing}{1.5}
\end{spacing}
\noindent \textbf{Breakdown Analysis.}
For a more detailed analysis of our method's performance across various query types, we segmented the execution results by query type and tabulated the statistics, as presented in Table \ref{breakdown_result}. 
From the table, it can be observed that the accuracy is relatively low for queries of Yes/No question type, Edge property type, and Attribute comparison type. These three types of queries all involve complex computational problems, so there may be some challenges in generating GQL with accurate calculation logic.

\begin{table}[]
\centering
\small
\begin{tabular*}{\hsize}{@{}@{\extracolsep{\fill}}l c c@{}}
\toprule
% \multicolumn{1}{l}{\multirow{2}{*}{\textbf{Method}}} & \multicolumn{2}{c}{\textbf{MediGQL}} \\
\multicolumn{1}{l}{\textbf{Query type}} &  \textbf{EM}       & \textbf{EX}    \\ 
\midrule
Entity property  & 81.87  & 80.63    \\ 
Numerical sorting  & 84.88  & 84.88    \\ 
Relationship inference  & 84.18  & 84.48    \\ 
Yes/No question  & 69.40  & 68.87    \\ 
Relationship filtering  & 79.55  & 77.27    \\ 
Attribute comparison  & 76.01  & 78.50    \\ 
Edge property  & 70.15  & 71.61    \\ 
String filtering  & 88.89  & 85.19    \\ 
\midrule
Total  & 79.65  & 73.75    \\ 
\bottomrule
\end{tabular*}
\caption{\label{breakdown_result}
Performance of our method on various types of queries in the FinGQL dataset.}
 \vspace{-1em}
\end{table}

\subsection{Ablation Study}
We conducted an ablation study to evaluate the effectiveness of components within our method, and the results are presented in Table~\ref{finetuned_method}.
The results indicate that removing the Chinese explanations of fields in the schema leads to a deterioration in performance, suggesting their contribution to addressing the Chinese NL2GQL task. 
Moreover, without any schema leads to a more significant performance drop, highlighting the beneficial role of schema as context in addressing the NL2GQL task.

\begin{table}[ht!]
\centering
\tabcolsep=0.2em
\small
% \begin{tabular}{lcc}
\begin{tabular*}{\hsize}{@{}@{\extracolsep{\fill}}l c c@{}}
\toprule
% \multicolumn{1}{l}{\multirow{2}{*}{\textbf{Method}}} & \multicolumn{2}{c}{\textbf{MediGQL}} \\
\multicolumn{1}{l}{\textbf{Method}} &     \textbf{EM}           & \textbf{EX}          \\ 
\midrule
Relevant Chinese Schema (Ours)                     & \textbf{86.05}        & \textbf{72.13}       \\ 
  \midrule
 % Relevant Schema     & 85.12 \textcolor{mylightgreen}{$\downarrow${(0.93)}} &  71.81 \textcolor{mylightgreen}{$\downarrow${(0.32)} }    \\
 %  Without  Schema   & 83.46 \textcolor{mylightgreen}{$\downarrow${(2.59)}} & 69.92  \textcolor{mylightgreen}{$\downarrow${(2.21)}}      \\ 
 %  Full Schema     & 85.29 \textcolor{mylightgreen}{$\downarrow${(0.76)}}      & 71.98  \textcolor{mylightgreen}{$\downarrow${(0.15)} }    \\
 %  Full Chinese Schema     & 85.55  \textcolor{mylightgreen}{$\downarrow${(0.50)}}      & 71.34  \textcolor{mylightgreen}{$\downarrow${(0.79)} }    \\
   Relevant Schema     & 85.12 \textcolor{mylightgreen}{$\downarrow${(0.9)}} &  71.81 \textcolor{mylightgreen}{$\downarrow${(0.3)} }    \\
  Without  Schema   & 83.46 \textcolor{mylightgreen}{$\downarrow${(2.6)}} & 69.92  \textcolor{mylightgreen}{$\downarrow${(2.2)}}      \\ 
  Full Schema     & 85.29 \textcolor{mylightgreen}{$\downarrow${(0.8)}}      & 71.98  \textcolor{mylightgreen}{$\downarrow${(0.1)} }    \\
  Full Chinese Schema     & 85.55  \textcolor{mylightgreen}{$\downarrow${(0.5)}}      & 71.34  \textcolor{mylightgreen}{$\downarrow${(0.8)} }    \\
\bottomrule
\end{tabular*}
\caption{\label{finetuned_method}
 Ablation study of our method on MediGQL. The green downward arrow denotes a decrease, and the green number in parentheses indicates the precise decrease value.}
 \vspace{-1.5em}
\end{table}

% \begin{table}[]
% \centering
% \small
% \resizebox{\linewidth}{!}{
% \begin{tabular}{cccccc}
% \toprule
% Number of F and O & \multicolumn{1}{c}{1} & 2 & 3 & 4 & 5  \\ \hline
% FinCQL            &               1000        & 1000  &  1000 & 1000  & 1000   \\
% MediCQL           &                1000       & 1000  & 1000  &  1000 &  1000  \\ 
% \hline
% Number of F and O & 6       & 7 & 8 & 9 & 10 \\ \hline
% FinCQL            &    1000             & 1000  &  1000 &  1000 &  1000  \\
% MediCQL           &      1000          & 1000  & 1000  & 1000  &  1000  \\ 
% \bottomrule
% \end{tabular}
% }
% \caption{\label{static_f_and_o}
% "F" represents functions, and "O" represents operators. The statistical count excludes instances of MATCH, WHERE, and RETURN.}
% \end{table}

\section{Conclusion}
\label{sec:conc}

In this paper, we present a pipeline for addressing the NL2GQL task on a given graph DB. Firstly, we construct an NL2GQL dataset based on the graph database. Then, we fine-tune LLMs using the constructed dataset to align LLMs with the graph database. Additionally, to further enhance the NL2GQL capability of LLMs, we propose a method that extracts schema relevant to NL as input context to guide LLM. Experimental results demonstrate that our approach significantly outperforms baseline methods.
\section*{Limitations}

Although our work is intended to be language-agnostic, it is regrettable that our experiments currently only involve Chinese data, as we only have access to domain graph data available in Chinese.

% 改进的地方
  % 1、related work 要和text-to-sql以及KBQA的比较
  % 2、主图需要添加一些解释
  % 3、方法名修改---two-instruction 要加一个相互验证的，并添加数据
  % 4、query type表格需要找一些复杂的例子，而且加一个数据集统计，多跳数据占比
  % 5、template 设计的合理，从实际场景提炼出来的，再加上数据验证
  % 6、NL-GQL的一致性验证
  % 7、detail retrieving examples in the In-Context Learning (ICL) baseline 
  % 8、Graph DB" and "graph database" 一致
  % 9、case study
  % 10、展示100条数据作为辅助了解，中文问题的英文翻译。
  % 11、schema linking的结果展示
\bibstyle{acl_natbib.bst}
\bibliography{custom,anthology}

\begin{thebibliography}{57}
\expandafter\ifx\csname natexlab\endcsname\relax\def\natexlab#1{#1}\fi

\bibitem[{Bai et~al.(2023)Bai, Bai, Chu, Cui, Dang, Deng, Fan, Ge, Han, Huang et~al.}]{bai2023qwen}
Jinze Bai, Shuai Bai, Yunfei Chu, Zeyu Cui, Kai Dang, Xiaodong Deng, Yang Fan, Wenbin Ge, Yu~Han, Fei Huang, et~al. 2023.
\newblock Qwen technical report.
\newblock \emph{arXiv preprint arXiv:2309.16609}.

\bibitem[{Besta et~al.(2023)Besta, Gerstenberger, Peter, Fischer, Podstawski, Barthels, Alonso, and Hoefler}]{besta:cs2023}
Maciej Besta, Robert Gerstenberger, Emanuel Peter, Marc Fischer, Micha{\l} Podstawski, Claude Barthels, Gustavo Alonso, and Torsten Hoefler. 2023.
\newblock Demystifying graph databases: Analysis and taxonomy of data organization, system designs, and graph queries.
\newblock \emph{ACM Computing Surveys}, 56(2):1--40.

\bibitem[{Chang and Fosler-Lussier(2023)}]{chang2023prompt}
Shuaichen Chang and Eric Fosler-Lussier. 2023.
\newblock How to prompt llms for text-to-sql: A study in zero-shot, single-domain, and cross-domain settings.
\newblock \emph{arXiv preprint arXiv:2305.11853}.

\bibitem[{Chen et~al.(2023{\natexlab{a}})Chen, Wang, Gao, Jiang, Chen, Zhang, Song, Xie, Kong, Li, Wan, Li, and Wang}]{chen2023huatuogptii}
Junying Chen, Xidong Wang, Anningzhe Gao, Feng Jiang, Shunian Chen, Hongbo Zhang, Dingjie Song, Wenya Xie, Chuyi Kong, Jianquan Li, Xiang Wan, Haizhou Li, and Benyou Wang. 2023{\natexlab{a}}.
\newblock Huatuogpt-ii, one-stage training for medical adaption of llms.

\bibitem[{Chen et~al.(2023{\natexlab{b}})Chen, Wang, Xing, huimin zheng, Xu, Fang, Wang, Li, Wu, Liu, and Xu}]{chen2023bianque}
Yirong Chen, Zhenyu Wang, Xiaofen Xing, huimin zheng, Zhipei Xu, Kai Fang, Junhong Wang, Sihang Li, Jieling Wu, Qi~Liu, and Xiangmin Xu. 2023{\natexlab{b}}.
\newblock Bianque: Balancing the questioning and suggestion ability of health llms with multi-turn health conversations polished by chatgpt.

\bibitem[{De~Cao et~al.(2021)De~Cao, Aziz, and Titov}]{de-cao-etal-2021-editing}
Nicola De~Cao, Wilker Aziz, and Ivan Titov. 2021.
\newblock \href {https://doi.org/10.18653/v1/2021.emnlp-main.522} {Editing factual knowledge in language models}.
\newblock In \emph{Proceedings of the 2021 Conference on Empirical Methods in Natural Language Processing}, pages 6491--6506, Online and Punta Cana, Dominican Republic. Association for Computational Linguistics.

\bibitem[{Diao et~al.(2023)Diao, Wang, Lin, and Zhang}]{diao2023active}
Shizhe Diao, Pengcheng Wang, Yong Lin, and Tong Zhang. 2023.
\newblock Active prompting with chain-of-thought for large language models.
\newblock \emph{arXiv preprint arXiv:2302.12246}.

\bibitem[{Dong et~al.(2023)Dong, Li, Dai, Zheng, Wu, Chang, Sun, Xu, Li, and Sui}]{dong2023survey}
Qingxiu Dong, Lei Li, Damai Dai, Ce~Zheng, Zhiyong Wu, Baobao Chang, Xu~Sun, Jingjing Xu, Lei Li, and Zhifang Sui. 2023.
\newblock \href {http://arxiv.org/abs/2301.00234} {A survey on in-context learning}.

\bibitem[{Ferrara(2023)}]{ferrara2023should}
Emilio Ferrara. 2023.
\newblock Should chatgpt be biased? challenges and risks of bias in large language models.
\newblock \emph{arXiv preprint arXiv:2304.03738}.

\bibitem[{Gao et~al.(2023{\natexlab{a}})Gao, Wang, Li, Sun, Qian, Ding, and Zhou}]{gao2023text}
Dawei Gao, Haibin Wang, Yaliang Li, Xiuyu Sun, Yichen Qian, Bolin Ding, and Jingren Zhou. 2023{\natexlab{a}}.
\newblock Text-to-sql empowered by large language models: A benchmark evaluation.
\newblock \emph{arXiv preprint arXiv:2308.15363}.

\bibitem[{Gao et~al.(2023{\natexlab{b}})Gao, Wang, Li, Sun, Qian, Ding, and Zhou}]{gao2023texttosql}
Dawei Gao, Haibin Wang, Yaliang Li, Xiuyu Sun, Yichen Qian, Bolin Ding, and Jingren Zhou. 2023{\natexlab{b}}.
\newblock \href {http://arxiv.org/abs/2308.15363} {Text-to-sql empowered by large language models: A benchmark evaluation}.

\bibitem[{Gao et~al.(2024)Gao, Xiong, Gao, Jia, Pan, Bi, Dai, Sun, Guo, Wang, and Wang}]{gao2024retrievalaugmented}
Yunfan Gao, Yun Xiong, Xinyu Gao, Kangxiang Jia, Jinliu Pan, Yuxi Bi, Yi~Dai, Jiawei Sun, Qianyu Guo, Meng Wang, and Haofen Wang. 2024.
\newblock \href {http://arxiv.org/abs/2312.10997} {Retrieval-augmented generation for large language models: A survey}.

\bibitem[{Gao et~al.(2023{\natexlab{c}})Gao, Xiong, Gao, Jia, Pan, Bi, Dai, Sun, and Wang}]{gao2023retrieval}
Yunfan Gao, Yun Xiong, Xinyu Gao, Kangxiang Jia, Jinliu Pan, Yuxi Bi, Yi~Dai, Jiawei Sun, and Haofen Wang. 2023{\natexlab{c}}.
\newblock Retrieval-augmented generation for large language models: A survey.
\newblock \emph{arXiv preprint arXiv:2312.10997}.

\bibitem[{Gu et~al.(2021)Gu, Kase, Vanni, Sadler, Liang, Yan, and Su}]{gu2021beyond}
Yu~Gu, Sue Kase, Michelle Vanni, Brian Sadler, Percy Liang, Xifeng Yan, and Yu~Su. 2021.
\newblock Beyond iid: three levels of generalization for question answering on knowledge bases.
\newblock In \emph{Proceedings of the Web Conference 2021}, pages 3477--3488.

\bibitem[{Gu et~al.(2023)Gu, Fan, Tang, Cao, Jia, Madden, and Du}]{gu2023few}
Zihui Gu, Ju~Fan, Nan Tang, Lei Cao, Bowen Jia, Sam Madden, and Xiaoyong Du. 2023.
\newblock Few-shot text-to-sql translation using structure and content prompt learning.
\newblock \emph{Proceedings of the ACM on Management of Data}, 1(2):1--28.

\bibitem[{Guo et~al.(2022)Guo, Li, Xiao, Tan, and Zhao}]{guo2022spcql}
Aibo Guo, Xinyi Li, Guanchen Xiao, Zhen Tan, and Xiang Zhao. 2022.
\newblock Spcql: A semantic parsing dataset for converting natural language into cypher.
\newblock In \emph{Proceedings of the 31st ACM International Conference on Information \& Knowledge Management}, pages 3973--3977.

\bibitem[{Hart et~al.(1968)Hart, Nilsson, and Raphael}]{hart1968formal}
Peter~E Hart, Nils~J Nilsson, and Bertram Raphael. 1968.
\newblock A formal basis for the heuristic determination of minimum cost paths.
\newblock \emph{IEEE transactions on Systems Science and Cybernetics}, 4(2):100--107.

\bibitem[{Hu et~al.(2021)Hu, Shen, Wallis, Allen-Zhu, Li, Wang, Wang, and Chen}]{hu2021lora}
Edward~J Hu, Yelong Shen, Phillip Wallis, Zeyuan Allen-Zhu, Yuanzhi Li, Shean Wang, Lu~Wang, and Weizhu Chen. 2021.
\newblock Lora: Low-rank adaptation of large language models.
\newblock \emph{arXiv preprint arXiv:2106.09685}.

\bibitem[{Jiang and Usbeck(2022)}]{jiang2022knowledg}
Longquan Jiang and Ricardo Usbeck. 2022.
\newblock \href {https://doi.org/10.1145/3477495.3531751} {Knowledge graph question answering datasets and their generalizability: Are they enough for future research?}
\newblock In \emph{Proceedings of the 45th International ACM SIGIR Conference on Research and Development in Information Retrieval}, SIGIR '22, page 3209–3218, New York, NY, USA. Association for Computing Machinery.

\bibitem[{Kaddour et~al.(2023)Kaddour, Harris, Mozes, Bradley, Raileanu, and McHardy}]{kaddour2023challenges}
Jean Kaddour, Joshua Harris, Maximilian Mozes, Herbie Bradley, Roberta Raileanu, and Robert McHardy. 2023.
\newblock Challenges and applications of large language models.
\newblock \emph{arXiv preprint arXiv:2307.10169}.

\bibitem[{Kang et~al.(2023{\natexlab{a}})Kang, Kwak, Baek, and Hwang}]{kang2023knowledge}
Minki Kang, Jin~Myung Kwak, Jinheon Baek, and Sung~Ju Hwang. 2023{\natexlab{a}}.
\newblock Knowledge graph-augmented language models for knowledge-grounded dialogue generation.
\newblock \emph{arXiv preprint arXiv:2305.18846}.

\bibitem[{Kang et~al.(2023{\natexlab{b}})Kang, Kwak, Baek, and Hwang}]{kang2023arxiv}
Minki Kang, Jin~Myung Kwak, Jinheon Baek, and Sung~Ju Hwang. 2023{\natexlab{b}}.
\newblock \href {http://arxiv.org/abs/2305.18846} {Knowledge graph-augmented language models for knowledge-grounded dialogue generation}.

\bibitem[{Kojima et~al.(2022)Kojima, Gu, Reid, Matsuo, and Iwasawa}]{kojima2022large}
Takeshi Kojima, Shixiang~Shane Gu, Machel Reid, Yutaka Matsuo, and Yusuke Iwasawa. 2022.
\newblock Large language models are zero-shot reasoners.
\newblock \emph{Advances in neural information processing systems}, 35:22199--22213.

\bibitem[{Li et~al.(2023)Li, Fan, Gu, Li, Duan, Dong, Liu, and Wang}]{li2023flexkbqa}
Zhenyu Li, Sunqi Fan, Yu~Gu, Xiuxing Li, Zhichao Duan, Bowen Dong, Ning Liu, and Jianyong Wang. 2023.
\newblock Flexkbqa: A flexible llm-powered framework for few-shot knowledge base question answering.
\newblock \emph{arXiv preprint arXiv:2308.12060}.

\bibitem[{Lin et~al.(2021)Lin, Tseng, and Byrne}]{lin-etal-2021-knowledge}
Weizhe Lin, Bo-Hsiang Tseng, and Bill Byrne. 2021.
\newblock \href {https://doi.org/10.18653/v1/2021.emnlp-main.620} {Knowledge-aware graph-enhanced {GPT}-2 for dialogue state tracking}.
\newblock In \emph{Proceedings of the 2021 Conference on Empirical Methods in Natural Language Processing}, pages 7871--7881, Online and Punta Cana, Dominican Republic. Association for Computational Linguistics.

\bibitem[{Liu et~al.(2022)Liu, Tam, Muqeeth, Mohta, Huang, Bansal, and Raffel}]{liu2022few}
Haokun Liu, Derek Tam, Mohammed Muqeeth, Jay Mohta, Tenghao Huang, Mohit Bansal, and Colin~A Raffel. 2022.
\newblock Few-shot parameter-efficient fine-tuning is better and cheaper than in-context learning.
\newblock \emph{Advances in Neural Information Processing Systems}, 35:1950--1965.

\bibitem[{Lopes et~al.(2023)Lopes, Rodrigues, Saraiva, Abbasi, Martins, and Wanzeller}]{lopes2023scalability}
Andr{\'e} Lopes, Diogo Rodrigues, Jo{\~a}o Saraiva, Maryam Abbasi, Pedro Martins, and Cristina Wanzeller. 2023.
\newblock Scalability and performance evaluation of graph database systems: A comparative study of neo4j, janusgraph, memgraph, nebulagraph, and tigergraph.
\newblock In \emph{2023 Second International Conference On Smart Technologies For Smart Nation (SmartTechCon)}, pages 537--542. IEEE.

\bibitem[{Lu et~al.(2022)Lu, Liu, Dai, Xiao, Lin, Han, Sun, and Wu}]{lu-etal-2022-unified}
Yaojie Lu, Qing Liu, Dai Dai, Xinyan Xiao, Hongyu Lin, Xianpei Han, Le~Sun, and Hua Wu. 2022.
\newblock \href {https://doi.org/10.18653/v1/2022.acl-long.395} {Unified structure generation for universal information extraction}.
\newblock In \emph{Proceedings of the 60th Annual Meeting of the Association for Computational Linguistics (Volume 1: Long Papers)}, pages 5755--5772, Dublin, Ireland. Association for Computational Linguistics.

\bibitem[{Luo et~al.(2023)Luo, Tang, Peng, Guo, Zhang, Ma, Dong, Song, Lin et~al.}]{luo2023chatkbqa}
Haoran Luo, Zichen Tang, Shiyao Peng, Yikai Guo, Wentai Zhang, Chenghao Ma, Guanting Dong, Meina Song, Wei Lin, et~al. 2023.
\newblock Chatkbqa: A generate-then-retrieve framework for knowledge base question answering with fine-tuned large language models.
\newblock \emph{arXiv preprint arXiv:2310.08975}.

\bibitem[{Meyer et~al.(2023)Meyer, Stadler, Frey, Radtke, Junghanns, Meissner, Dziwis, Bulert, and Martin}]{meyer2023llm}
Lars-Peter Meyer, Claus Stadler, Johannes Frey, Norman Radtke, Kurt Junghanns, Roy Meissner, Gordian Dziwis, Kirill Bulert, and Michael Martin. 2023.
\newblock Llm-assisted knowledge graph engineering: Experiments with chatgpt.
\newblock \emph{arXiv preprint arXiv:2307.06917}.

\bibitem[{Min et~al.(2023)Min, Ross, Sulem, Veyseh, Nguyen, Sainz, Agirre, Heintz, and Roth}]{min2023recent}
Bonan Min, Hayley Ross, Elior Sulem, Amir Pouran~Ben Veyseh, Thien~Huu Nguyen, Oscar Sainz, Eneko Agirre, Ilana Heintz, and Dan Roth. 2023.
\newblock Recent advances in natural language processing via large pre-trained language models: A survey.
\newblock \emph{ACM Computing Surveys}, 56(2):1--40.

\bibitem[{Modarressi et~al.(2023)Modarressi, Imani, Fayyaz, and Sch{\"u}tze}]{modarressi2023ret}
Ali Modarressi, Ayyoob Imani, Mohsen Fayyaz, and Hinrich Sch{\"u}tze. 2023.
\newblock Ret-llm: Towards a general read-write memory for large language models.
\newblock \emph{arXiv preprint arXiv:2305.14322}.

\bibitem[{Monteiro et~al.(2023)Monteiro, S{\'a}, and Bernardino}]{monteiro2023experimental}
J{\'e}ssica Monteiro, Filipe S{\'a}, and Jorge Bernardino. 2023.
\newblock Experimental evaluation of graph databases: Janusgraph, nebula graph, neo4j, and tigergraph.
\newblock \emph{Applied Sciences}, 13(9):5770.

\bibitem[{Nan et~al.(2023)Nan, Zhao, Zou, Ri, Tae, Zhang, Cohan, and Radev}]{nan2023enhancing}
Linyong Nan, Yilun Zhao, Weijin Zou, Narutatsu Ri, Jaesung Tae, Ellen Zhang, Arman Cohan, and Dragomir Radev. 2023.
\newblock Enhancing text-to-{SQL} capabilities of large language models: A study on prompt design strategies.
\newblock In \emph{The 2023 Conference on Empirical Methods in Natural Language Processing}.

\bibitem[{Pan(2009)}]{pan2009resource}
Jeff~Z Pan. 2009.
\newblock Resource description framework.
\newblock In \emph{Handbook on ontologies}, pages 71--90. Springer.

\bibitem[{Pan et~al.(2024)Pan, Luo, Wang, Chen, Wang, and Wu}]{pan2024unifying}
Shirui Pan, Linhao Luo, Yufei Wang, Chen Chen, Jiapu Wang, and Xindong Wu. 2024.
\newblock Unifying large language models and knowledge graphs: A roadmap.
\newblock \emph{IEEE Transactions on Knowledge and Data Engineering}.

\bibitem[{Pourreza and Rafiei(2023)}]{pourreza2023dinsql}
Mohammadreza Pourreza and Davood Rafiei. 2023.
\newblock \href {http://arxiv.org/abs/2304.11015} {Din-sql: Decomposed in-context learning of text-to-sql with self-correction}.

\bibitem[{Saxena et~al.(2020)Saxena, Tripathi, and Talukdar}]{saxena-etal-2020-improving}
Apoorv Saxena, Aditay Tripathi, and Partha Talukdar. 2020.
\newblock \href {https://doi.org/10.18653/v1/2020.acl-main.412} {Improving multi-hop question answering over knowledge graphs using knowledge base embeddings}.
\newblock In \emph{Proceedings of the 58th Annual Meeting of the Association for Computational Linguistics}, pages 4498--4507, Online. Association for Computational Linguistics.

\bibitem[{Sen et~al.(2023)Sen, Mavadia, and Saffari}]{sen:nlrse2023}
Priyanka Sen, Sandeep Mavadia, and Amir Saffari. 2023.
\newblock Knowledge graph-augmented language models for complex question answering.
\newblock In \emph{Proceedings of the 1st Workshop on Natural Language Reasoning and Structured Explanations (NLRSE)}.

\bibitem[{Tang et~al.(2020)Tang, Tran, Li, Chen, Goyal, Chaudhary, Gu, and Fan}]{tang2020multilingual}
Yuqing Tang, Chau Tran, Xian Li, Peng-Jen Chen, Naman Goyal, Vishrav Chaudhary, Jiatao Gu, and Angela Fan. 2020.
\newblock \href {http://arxiv.org/abs/2008.00401} {Multilingual translation with extensible multilingual pretraining and finetuning}.

\bibitem[{Wang et~al.(2020)Wang, Yang, Zhang, and Lin}]{wang2020empirical}
Ran Wang, Zhengyi Yang, Wenjie Zhang, and Xuemin Lin. 2020.
\newblock An empirical study on recent graph database systems.
\newblock In \emph{Knowledge Science, Engineering and Management: 13th International Conference, KSEM 2020, Hangzhou, China, August 28--30, 2020, Proceedings, Part I 13}, pages 328--340. Springer.

\bibitem[{Wang et~al.(2023)Wang, Yang, Qiu, Liang, He, Gu, Xiao, and Wang}]{wang2023knowledgpt}
Xintao Wang, Qianwen Yang, Yongting Qiu, Jiaqing Liang, Qianyu He, Zhouhong Gu, Yanghua Xiao, and Wei Wang. 2023.
\newblock Knowledgpt: Enhancing large language models with retrieval and storage access on knowledge bases.
\newblock \emph{arXiv preprint arXiv:2308.11761}.

\bibitem[{Wang et~al.(2022)Wang, Kordi, Mishra, Liu, Smith, Khashabi, and Hajishirzi}]{wang2022self}
Yizhong Wang, Yeganeh Kordi, Swaroop Mishra, Alisa Liu, Noah~A Smith, Daniel Khashabi, and Hannaneh Hajishirzi. 2022.
\newblock Self-instruct: Aligning language model with self generated instructions.
\newblock \emph{arXiv preprint arXiv:2212.10560}.

\bibitem[{Wei et~al.(2022{\natexlab{a}})Wei, Tay, Bommasani, Raffel, Zoph, Borgeaud, Yogatama, Bosma, Zhou, Metzler et~al.}]{wei2022emergent}
Jason Wei, Yi~Tay, Rishi Bommasani, Colin Raffel, Barret Zoph, Sebastian Borgeaud, Dani Yogatama, Maarten Bosma, Denny Zhou, Donald Metzler, et~al. 2022{\natexlab{a}}.
\newblock Emergent abilities of large language models.
\newblock \emph{arXiv preprint arXiv:2206.07682}.

\bibitem[{Wei et~al.(2022{\natexlab{b}})Wei, Wang, Schuurmans, Bosma, ichter, Xia, Chi, Le, and Zhou}]{NEURIPS2022_9d560961}
Jason Wei, Xuezhi Wang, Dale Schuurmans, Maarten Bosma, brian ichter, Fei Xia, Ed~Chi, Quoc~V Le, and Denny Zhou. 2022{\natexlab{b}}.
\newblock \href {https://proceedings.neurips.cc/paper_files/paper/2022/file/9d5609613524ecf4f15af0f7b31abca4-Paper-Conference.pdf} {Chain-of-thought prompting elicits reasoning in large language models}.
\newblock In \emph{Advances in Neural Information Processing Systems}, volume~35, pages 24824--24837. Curran Associates, Inc.

\bibitem[{Wei et~al.(2022{\natexlab{c}})Wei, Wang, Schuurmans, Bosma, Xia, Chi, Le, Zhou et~al.}]{wei2022chain}
Jason Wei, Xuezhi Wang, Dale Schuurmans, Maarten Bosma, Fei Xia, Ed~Chi, Quoc~V Le, Denny Zhou, et~al. 2022{\natexlab{c}}.
\newblock Chain-of-thought prompting elicits reasoning in large language models.
\newblock \emph{Advances in Neural Information Processing Systems}, 35:24824--24837.

\bibitem[{Xi et~al.(2023)Xi, Chen, Guo, He, Ding, Hong, Zhang, Wang, Jin, Zhou et~al.}]{xi2023rise}
Zhiheng Xi, Wenxiang Chen, Xin Guo, Wei He, Yiwen Ding, Boyang Hong, Ming Zhang, Junzhe Wang, Senjie Jin, Enyu Zhou, et~al. 2023.
\newblock The rise and potential of large language model based agents: A survey.
\newblock \emph{arXiv preprint arXiv:2309.07864}.

\bibitem[{Xue et~al.(2021)Xue, Constant, Roberts, Kale, Al-Rfou, Siddhant, Barua, and Raffel}]{xue-etal-2021-mt5}
Linting Xue, Noah Constant, Adam Roberts, Mihir Kale, Rami Al-Rfou, Aditya Siddhant, Aditya Barua, and Colin Raffel. 2021.
\newblock \href {https://doi.org/10.18653/v1/2021.naacl-main.41} {m{T}5: A massively multilingual pre-trained text-to-text transformer}.
\newblock In \emph{Proceedings of the 2021 Conference of the North American Chapter of the Association for Computational Linguistics: Human Language Technologies}, pages 483--498, Online. Association for Computational Linguistics.

\bibitem[{Yang et~al.(2023{\natexlab{a}})Yang, Xiao, Wang, Zhang, Bian, Yin, Lv, Pan, Wang, Yan et~al.}]{yang2023baichuan}
Aiyuan Yang, Bin Xiao, Bingning Wang, Borong Zhang, Ce~Bian, Chao Yin, Chenxu Lv, Da~Pan, Dian Wang, Dong Yan, et~al. 2023{\natexlab{a}}.
\newblock Baichuan 2: Open large-scale language models.
\newblock \emph{arXiv preprint arXiv:2309.10305}.

\bibitem[{Yang et~al.(2023{\natexlab{b}})Yang, Teng, Dong, and Bo}]{yang2023llm}
Shuangtao Yang, Mao Teng, Xiaozheng Dong, and Fu~Bo. 2023{\natexlab{b}}.
\newblock Llm-based sparql generation with selected schema from large scale knowledge base.
\newblock In \emph{China Conference on Knowledge Graph and Semantic Computing}, pages 304--316. Springer.

\bibitem[{Yu et~al.(2018)Yu, Zhang, Yang, Yasunaga, Wang, Li, Ma, Li, Yao, Roman et~al.}]{yu2018spider}
Tao Yu, Rui Zhang, Kai Yang, Michihiro Yasunaga, Dongxu Wang, Zifan Li, James Ma, Irene Li, Qingning Yao, Shanelle Roman, et~al. 2018.
\newblock Spider: A large-scale human-labeled dataset for complex and cross-domain semantic parsing and text-to-sql task.
\newblock \emph{arXiv preprint arXiv:1809.08887}.

\bibitem[{Zeng et~al.(2022)Zeng, Liu, Du, Wang, Lai, Ding, Yang, Xu, Zheng, Xia et~al.}]{zeng2022glm}
Aohan Zeng, Xiao Liu, Zhengxiao Du, Zihan Wang, Hanyu Lai, Ming Ding, Zhuoyi Yang, Yifan Xu, Wendi Zheng, Xiao Xia, et~al. 2022.
\newblock Glm-130b: An open bilingual pre-trained model.
\newblock \emph{arXiv preprint arXiv:2210.02414}.

\bibitem[{Zhao et~al.(2023)Zhao, Zhou, Li, Tang, Wang, Hou, Min, Zhang, Zhang, Dong et~al.}]{zhao2023survey}
Wayne~Xin Zhao, Kun Zhou, Junyi Li, Tianyi Tang, Xiaolei Wang, Yupeng Hou, Yingqian Min, Beichen Zhang, Junjie Zhang, Zican Dong, et~al. 2023.
\newblock A survey of large language models.
\newblock \emph{arXiv preprint arXiv:2303.18223}.

\bibitem[{Zhao et~al.(2024)Zhao, Liu, French, and Stewart}]{zhao_cySpider}
Ziyu Zhao, Wei Liu, Tim French, and Michael Stewart. 2024.
\newblock Cyspider: A neural semantic parsing corpus with baseline models for property graphs.
\newblock In \emph{AI 2023: Advances in Artificial Intelligence}, pages 120--132, Singapore. Springer Nature Singapore.

\bibitem[{Zhao et~al.(2022)Zhao, Stewart, Liu, French, and Hodkiewicz}]{zhao2022natural}
Ziyu Zhao, Michael Stewart, Wei Liu, Tim French, and Melinda Hodkiewicz. 2022.
\newblock Natural language query for technical knowledge graph navigation.
\newblock In \emph{Australasian Conference on Data Mining}, pages 176--191. Springer.

\bibitem[{Zhou et~al.(2022)Zhou, Sch{\"a}rli, Hou, Wei, Scales, Wang, Schuurmans, Cui, Bousquet, Le et~al.}]{zhou2022least}
Denny Zhou, Nathanael Sch{\"a}rli, Le~Hou, Jason Wei, Nathan Scales, Xuezhi Wang, Dale Schuurmans, Claire Cui, Olivier Bousquet, Quoc Le, et~al. 2022.
\newblock Least-to-most prompting enables complex reasoning in large language models.
\newblock \emph{arXiv preprint arXiv:2205.10625}.

\bibitem[{Zhou et~al.(2023)Zhou, Yu, Tian, Chen, Zhou, Yu, Ji, Liu, Ye, and Chai}]{zhou2023r3nl2gql}
Yuhang Zhou, He~Yu, Siyu Tian, Dan Chen, Liuzhi Zhou, Xinlin Yu, Chuanjun Ji, Sen Liu, Guangnan Ye, and Hongfeng Chai. 2023.
\newblock $r^3$-nl2gql: A hybrid models approach for for accuracy enhancing and hallucinations mitigation.

\end{thebibliography}

\appendix
\newpage

\appendix

\section{Query Type}
\label{query_type_show}

Based on the logical structure between nodes, edges, and properties that may be involved in queries, we summarized the queries and classified them into 8 query types, which are presented in Table \ref{table:question_types}.

\section{Dataset Presentation}
\label{dataset_detail_show}

The \textbf{FinGQL} dataset comprises a total of 10,000 records, each consisting of an \textit{\textbf{NL}} along with its corresponding \textit{\textbf{GQL}}, a \textit{\textbf{template NL}} along with its corresponding \textit{\textbf{template GQL}}, the corresponding GQL \textit{\textbf{Answer}}, and the \textit{\textbf{Query Type ID}}.
Furthermore, \textit{\textbf{Nodes}} and \textit{\textbf{Edges}} are extracted from the GQL to provide Relevant Schema during training.
Similarly, the \textbf{MedGQL} dataset consists of 8,550 records.

An example from the FinGQL dataset is as follows:
\begin{itemize}
    \item \textit{\textbf{NL}}: \\
    {\small \begin{CJK*}{UTF8}{gbsn} 钢铁行业中，有哪些股票的开盘价大于1？\end{CJK*} \\
    \textit{(In the steel industry, which stocks have opening prices greater than 1?)}}

    \item \textit{\textbf{GQL}}: \\ 
    {\small MATCH (t:trade\{name:"\begin{CJK*}{UTF8}{gbsn}钢铁\end{CJK*}"\textit{(steel)}\}) \\
    <-[bt:belong\_to]-(s:stock)-[hsd:has\_stock\_data]->(sd:stock\_data) WHERE sd.stock\_data.opening\_price > 1 RETURN s.stock.name}

    \item \textit{\textbf{template NL}}: \\
    {\small \begin{CJK*}{UTF8}{gbsn} [t]行业中，有哪些股票的开盘价大于[m]？\end{CJK*} \\
    \textit{(In the [t] industry, which stocks have opening prices greater than [m]?)}}

    \item \textit{\textbf{template GQL}}: \\
    {\small MATCH (t:trade\{name:'[t]'\}) \\
    <-[bt:belong\_to]-(s:stock)-[hsd:has\_stock\_data]->(sd:stock\_data) WHERE sd.stock\_data.opening\_price > [m] RETURN s.stock.name}

    \item \textit{\textbf{Query Type ID}}: 
    {\small 6}

    \item \textit{\textbf{Nodes}}: 
    {\small ["stock", "stock\_data"]}

    \item \textit{\textbf{Edges}}: 
    {\small ["has\_stock\_data", "belong\_to"]}
\end{itemize}

% \textit{"NL":} \begin{CJK*}{UTF8}{gbsn} 钢铁行业中，有哪些股票的开盘价大于1？\end{CJK*}
% (In the steel industry, which stocks have opening prices greater than 1?)

% \textbf{"GQL":} MATCH (t:trade\{name:\begin{CJK*}{UTF8}{gbsn}钢铁\end{CJK*}\})<-[bt:belong\_to]-(s:stock)-[hsd:has\_stock\_data]->(sd:stock\_data) WHERE sd.stock\_data.opening\_price > [m] RETURN s.stock.name

% \textbf{"template NL":} \begin{CJK*}{UTF8}{gbsn} [t]行业中，有哪些股票的开盘价大于[m]？\end{CJK*}
% (In the [t] industry, which stocks have opening prices greater than [m]?)

% \textbf{"template GQL":} MATCH (t:trade\{name:'[t]'\})<-[bt:belong\_to]-(s:stock)-[hsd:has\_stock\_data]->(sd:stock\_data) WHERE sd.stock\_data.opening\_price > [m] RETURN s.stock.name

% \textbf{"Query Type ID":} 6

% \textbf{"Nodes":} ["stock", "stock\_data"]

% \textbf{"Edges":} ["has\_stock\_data", "belong\_to"]

% \begin{verbatim}
% {
%   "NL": "\begin{CJK*}{UTF8}{gbsn} 钢铁行业中，有哪些股票的开盘价大于1？\end{CJK*}",
%   "GQL": \begin{CJK*}{UTF8}{gbsn}MATCH (t:trade{name:'钢铁'})<-[bt:belong_to]-(s:stock)-[hsd:has_stock_data]->(sd:stock_data) WHERE sd.stock_data.opening_price > 1 RETURN s.stock.name \end{CJK*},
%   "template NL": ""[t]行业中，有哪些股票的开盘价大于[m]？"",
%   "template GQL": MATCH (t:trade{name:'[t]'})<-[bt:belong_to]-(s:stock)-[hsd:has_stock_data]->(sd:stock_data) WHERE sd.stock_data.opening_price > [m] RETURN s.stock.name,
%   "Query Type ID": 6,
%   "Nodes": ["stock", "stock_data"]
%   "Edges": ["has_stock_data", "belong_to"]
% }
% \end{verbatim}

\section{NL-GQL Data Pairs Generation Task Description }
\label{task_desc}
We aim to guide ChatGPT in generating NL-GQL template data pairs of the desired query type. Merely including schema and demonstrations in the prompt is not sufficient. Detailed descriptions of the desired query types and explanations of the specific constraints associated with each type are also necessary. Our task description follows the format below:

``\textit{
Please generate a new NL-GQL template data pair of the same type as the provided NL based on the given Graph database schema and NL-GQL demonstrations.
In the Graph database schema section, Nodes contain information about all entities, including node tags and their corresponding properties. Edges contain information about all relationships, including the head and tail nodes of the edge and any edge properties.
In the NL-GQL demonstrations, named entities are replaced with placeholders. For example, 'stock' is represented by '[s]', 'chairman' by '[c]', 'trade' by '[t]'...(Description of all replacement characters for nodes and edges in the schema is provided below. Due to space constraints, only a portion is written.) The query type is \{ query type \}, in this type \{ query type description\}''}.

In the above task description, replace "\{query type\}" and its corresponding "\{query type description\}" with the type of NL to be generated.
The query type descriptions we summarized are shown in Table \ref{table:query_type_desc}.

% Based on the provided schema of the Graph database and NL-GQL example, please generate NL-GQL data. The NL type is xxx, which pertains to...
% describing the query type of

\section{Ablation Study on FinGQL}
We present Table \ref{ablation_FinGQL} to showcase ablation studies on FinGQL datasets. Additionally, we recognize the importance of relevant Chinese schema in assisting LLMs to generate more accurate GQL. It is worth noting that we did not conduct experiments with the full schema approach due to the 2048 input length limit of Qwen-14B-Chat. Adding the full schema would exceed this limit. This underscores the importance of the strategy to extract the relevant schema of NL in our method.

\begin{table}[th!]
\centering
\tabcolsep=0.3em
\small
\begin{tabular}{lcc}
\toprule
% \multicolumn{1}{l}{\multirow{2}{*}{\textbf{Method}}} & \multicolumn{2}{c}{\textbf{FinGQL}} \\
% \cmidrule{2-3}
\multicolumn{1}{c}{\textbf{Method}}                        & \textbf{EM}           & \textbf{EX}          \\ 
\midrule
Relevant Chinese Schema                      & 79.65        & 73.75       \\ 
  \midrule
 Relevant Schema     & 77.85 \textcolor{mylightgreen}{$\downarrow${(1.8)}}      & 71.85  \textcolor{mylightgreen}{$\downarrow${(1.9)} }    \\
  Without  Schema   & 77.35 \textcolor{mylightgreen}{$\downarrow${(2.3)}}         & 71.55  \textcolor{mylightgreen}{$\downarrow${(2.2)}}      \\ 
\bottomrule
\end{tabular}
\caption{\label{ablation_FinGQL}
 Ablation study of our method on FinGQL. The green downward arrow denotes a decrease, and the green number in parentheses indicates the precise decrease value.}
 \vspace{-1em}
\end{table}

\begin{table*}[ht!]
\centering  
\small
\begin{tabular}{l|cc}
\toprule
\textbf{Query type}   & \textbf{Type\#1 Entity property}                                                                                                                                                                                                & \textbf{Type\#2 Numerical sorting}                                                                                                                                                                                                                       \\ \midrule
\textbf{NL template}  & What is {[}entity{]}’s {[}property{]}?                                                                                                                                                                               & \begin{tabular}[c]{@{}c@{}}The {[}property2{]} of node with the highest \\ {[}property1{]}?\end{tabular}                                                                                                                                     \\ \midrule
\textbf{GQL template} & \begin{tabular}[c]{@{}c@{}}MATCH (s:node\{name:'{[}entity{]}'\}) \\ RETURN s.node.{[}property{]}\end{tabular}                                                                                                        & \begin{tabular}[c]{@{}c@{}}MATCH (s:node)  WITH s.node.{[}property2{]} AS n1, \\ s.node.{[}property1{]} AS n2 ORDER BY n2 DESC \\ LIMIT 1 RETURN n1\end{tabular}                                                                             \\ \midrule \midrule
\textbf{Query type}   & \textbf{Type\#3 Relationship inference}                                                                                                                                                                                          & \textbf{Type\#4 Yes/No question}                                                                                                                                                                                                                         \\ \midrule
\textbf{NL template}  & The {[}property{]} of node1 rel2 by {[}entity{]}?                                                                                                                                                                    & \begin{tabular}[c]{@{}c@{}}Is the {[}property{]} of the {[}entity{]} \\  greater than {[}mount{]}?\end{tabular}                                                                                                                              \\ \midrule
\textbf{GQL template} & \begin{tabular}[c]{@{}c@{}}MATCH (s1:node1)\textless{}-{[}:rel1{]}-(s2:node2)\\ \textless{}-{[}:rel2{]}-(s3:node3\{name: '{[}entity{]}'\}) \\ RETURN s1.node1.{[}property{]}\end{tabular}                            & \begin{tabular}[c]{@{}c@{}}MATCH (s:node\{name: '{[}entity{]}'\}) \\ RETURN s.node.{[}property{]} \textgreater {[}mount{]}\end{tabular}                                                                                                      \\ \midrule \midrule
\textbf{Query type}   & \textbf{Type\#5 Relationship filtering}                                                                                                                                                                                          & \textbf{Type\#6 Attribute comparison}                                                                                                                                                                                                                    \\ \midrule
\textbf{NL template}  & \begin{tabular}[c]{@{}c@{}}The {[}property{]} of node2  that rel {[}entity1{]} \\ and {[}entity2{]}?\end{tabular}                                                                                                    & \begin{tabular}[c]{@{}c@{}}How much does the {[}property{]} of {[}entity1{]} \\ differ from that of {[}entity2{]}?\end{tabular}                                                                                                              \\ \midrule
\textbf{GQL template} & \begin{tabular}[c]{@{}c@{}}MATCH (s1:node1\{name: '{[}entity1{]}'\})\textless{}-{[}:rel{]}\\ -(s2:node2)-{[}:rel{]}-\textgreater{}(s3:node1\{name: '{[}entity2{]}'\}) \\ RETURN s2:node2.{[}property{]}\end{tabular} & \begin{tabular}[c]{@{}c@{}}MATCH (s1:node\{name: '{[}entity1{]}'\}) \\ WITH s1.node.{[}property{]} AS t1 \\ MATCH (s2:node\{name: '{[}entity2{]}'\}) \\ WITH ABS(s2.node.{[}property{]} - t1) \\ as abs\_diff  RETURN abs\_diff\end{tabular} \\ \midrule \midrule
\textbf{Query type}   & \textbf{Type\#7 Edge property}                                                                                                                                                                                                   & \textbf{Type\#8 String filtering}                                                                                                                                                                                                                        \\ \midrule
\textbf{NL template}  & \begin{tabular}[c]{@{}c@{}}The {[}property{]} of node2  that rel {[}entity{]}  with \\ {[}r\_property{]} less than {[}m{]}?\end{tabular}                                                                             & \begin{tabular}[c]{@{}c@{}}The {[}property{]} of the node containing \\ the {[}string{]}?\end{tabular}                                                                                                                                       \\ \midrule
\textbf{GQL template} & \begin{tabular}[c]{@{}c@{}}MATCH (s1:node1\{name: '{[}entity{]}'\})-{[}r:rel{]}\\ -\textgreater{}(s2:node2) WHERE r.{[}r\_property{]} \textless {[}m{]} \\ RETURN s2.node2.{[}property{]}\end{tabular}               & \begin{tabular}[c]{@{}c@{}}MATCH (s:node) WHERE s.node.{[}property{]}  \\ CONTAINS  '{[}string{]}' RETURN s.node.{[}property{]}\end{tabular}                                                                                                 \\ \bottomrule
\end{tabular}
\caption{8 canonical query types along with their corresponding NL and GQL templates.}
% \yscomment{Please re-organize the type by giving a type followed by an example (a question template and a cypher template).}}
\label{table:question_types}
\vspace{-1em}
\end{table*}

\begin{table*}[]
\centering
\tabcolsep=0.3em
\small
\begin{tabular}{ll}
\toprule
\textbf{Query type}                     & \textbf{Query type descriptions} \\
\midrule
\textbf{Type\#1 Entity property}        &   \makecell[l]{The generated NL and GQL are both related to simple property queries of nodes and \\do not involve any computations.} \\
\midrule
\textbf{Type\#2 Numerical sorting}      &     \makecell[l]{The generated NL and GQL are both related to sorting concepts and numerical values.}\\ 
\midrule
\textbf{Type\#3 Relationship inference} &     \makecell[l]{The generated NL and GQL are both related to reasoning about relationships between\\ nodes and edges.}                    \\
\midrule
\textbf{Type\#4 Yes/No question}        &      \makecell[l]{The generated NL and GQL both pertain to 'whether' or 'existence' related queries.}\\
\midrule
\textbf{Type\#5 Relationship filtering} &    \makecell[l]{The generated NL and GQL are both related to filtering relationships on multi-hop\\subgraphs associated with nodes.}   \\
\midrule
\textbf{Type\#6 Attribute comparison}   &  \makecell[l]{The generated NL and GQL are both related to aggregate calculations of node\\properties.}                       \\
\midrule
\textbf{Type\#7 Edge property}          &   \makecell[l]{The generated NL and GQL are both related to comparing and aggregating\\calculations of edge properties.}                      \\
\midrule
\textbf{Type\#8 String filtering}       &   \makecell[l]{The generated NL and GQL are both related to string matching.}                     \\ \bottomrule
\end{tabular}
\caption{Description of each query type.}
% \yscomment{Please re-organize the type by giving a type followed by an example (a question template and a cypher template).}}
\label{table:query_type_desc}
\end{table*}

\label{sec:appendix}

\end{document}